\documentclass{article}

\usepackage{amsmath,amssymb,amsthm}
\usepackage{algorithmic}
\usepackage{algorithm}
\usepackage{array}
\usepackage[caption=false,font=normalsize,labelfont=sf,textfont=sf]{subfig}
\usepackage{textcomp}
\usepackage{stfloats}
\usepackage{url}
\usepackage{verbatim}
\usepackage{graphicx}
\usepackage{cite}
\usepackage{enumitem}
\usepackage{multirow}
\usepackage{svg}

\usepackage{arxiv}

\usepackage[utf8]{inputenc} 
\usepackage[T1]{fontenc}    
\usepackage{hyperref}       
\usepackage{url}            
\usepackage{booktabs}       
\usepackage{amsfonts}       
\usepackage{nicefrac}       
\usepackage{microtype}      
\usepackage{lipsum}		
\usepackage{graphicx}
\usepackage{doi}

\usepackage{tikz}
\usepackage{textcomp}
\newcommand\copyrighttext{%
\footnotesize \textcopyright 2023 IEEE. Personal use of this material is permitted.
Permission from IEEE must be obtained for all other uses, in any current or future
media, including reprinting/republishing this material for advertising or promotional
purposes, creating new collective works, for resale or redistribution to servers or
lists, or reuse of any copyrighted component of this work in other works.

\doi{10.1109/TAES.2023.3270111}}
\newcommand\copyrightnotice{%
\begin{tikzpicture}[remember picture,overlay]
\node[anchor=south,yshift=20pt] at (current page.south) {\fbox{\parbox{\dimexpr\textwidth-\fboxsep-\fboxrule\relax}{\copyrighttext}}};
\end{tikzpicture}%
}

\title{Deep Learning-Based Multiband Signal Fusion for 3-D SAR Super-Resolution}

\author{ \href{https://orcid.org/0000-0002-3388-4805}{\includegraphics[scale=0.06]{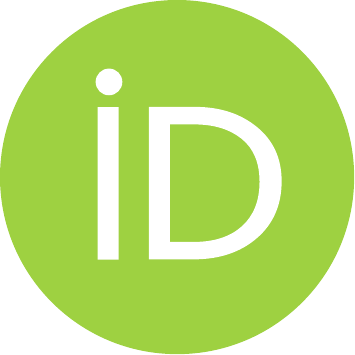}\hspace{1mm}Josiah W. Smith} \\
	Department of Electrical and Computer Engineering\\
	The University of Texas at Dallas\\
	Richardson, TX 75080 \\
	\texttt{josiah.smith@utdallas.edu} \\
	\And
	\href{https://orcid.org/0000-0001-7229-1765}{\includegraphics[scale=0.06]{orcid.pdf}\hspace{1mm}Murat Torlak}\thanks{The work of Murat Torlak (while serving at NSF) was supported by NSF.} \\
	Department of Electrical and Computer Engineering\\
	The University of Texas at Dallas\\
	Richardson, TX 75080 \\
	\texttt{torlak@utdallas.edu} \\
}

\date{}

\renewcommand{\headeright}{\textit{IEEE Trans. Aerosp. Electron. Syst.} (Accepted)}
\renewcommand{\undertitle}{\textit{IEEE Trans. Aerosp. Electron. Syst.} (Accepted)}
\renewcommand{\shorttitle}{Multiband Fusion}

\hypersetup{
pdftitle={Deep Learning-Based Multiband Signal Fusion for 3-D SAR Super-Resolution},
pdfsubject={cs.CV, eess.SP, cs.AI},
pdfauthor={Josiah W.~Smith, Murat Torlak},
pdfkeywords={deep neural network, millimeter-wave (mmWave), multiband signal, radar resolution, signal fusion, synthetic aperture radar (SAR)},
}

\begin{document}
\maketitle
\copyrightnotice

\begin{abstract}
Three-dimensional (3-D) synthetic aperture radar (SAR) is widely used in many security and industrial applications requiring high-resolution imaging of concealed or occluded objects. 
The ability to resolve intricate 3-D targets is essential to the performance of such applications and depends directly on system bandwidth. 
However, because high-bandwidth systems face several prohibitive hurdles, an alternative solution is to operate multiple radars at distinct frequency bands and fuse the multiband signals. 
Current multiband signal fusion methods assume a simple target model and a small number of point reflectors, which is invalid for realistic security screening and industrial imaging scenarios wherein the target model effectively consists of a large number of reflectors. 
To the best of our knowledge, this study presents the first use of deep learning for multiband signal fusion. 
The proposed network, called $kR$-Net, employs a hybrid, dual-domain complex-valued convolutional neural network (CV-CNN) to fuse multiband signals and impute the missing samples in the frequency gaps between subbands. 
By exploiting the relationships in both the wavenumber domain and wavenumber spectral domain, the proposed framework overcomes the drawbacks of existing multiband imaging techniques for realistic scenarios at a fraction of the computation time of existing multiband fusion algorithms. 
Our method achieves high-resolution imaging of intricate targets previously impossible using conventional techniques and enables finer resolution capacity for concealed weapon detection and occluded object classification using multiband signaling without requiring more advanced hardware. 
Furthermore, a fully integrated multiband imaging system is developed using commercially available millimeter-wave (mmWave) radars for efficient multiband imaging. 
Using two mmWave radars, each with a bandwidth of 4 GHz operating at 60 GHz and 77 GHz, the proposed $kR$-Net is employed to achieve an effective bandwidth of 21 GHz by robustly estimating the full-band signal. 
Additionally, the generalizability of the proposed technique is demonstrated across multiband sensing scenarios in the mmWave and terahertz (THz) frequencies. 
Extensive numerical simulations and empirical experiments are conducted to illustrate the superiority of our approach over existing methods for a diverse set of realistic 3-D SAR imaging scenarios. 
\end{abstract}


\section{Introduction}
\label{sec:dri_intro}
N{\scshape ear-field} synthetic aperture radar (SAR) imaging has received increasing attention for its use in nondestructive testing, concealed weapon detection, medical imaging, and remote sensing. 
Owing to the non-ionizing nature of electromagnetic (EM) waves at millimeter-wave (mmWave) and terahertz (THz) frequencies, they are considered safe for human applications. 
Consequently, mmWave devices have been used for many sensing and imaging applications.
Image resolution is a key characteristic of near-field SAR imaging.
In particular, because the downrange resolution is inversely proportional to the system bandwidth, ultrawideband transceivers are continually challenged to achieve greater bandwidths because gigahertz-scale bandwidths are required to achieve cm and sub-cm resolutions \cite{yanik2020development}.
Although sophisticated lab equipment can fulfill the bandwidth demands of certain applications \cite{batra2021short}, end-user applications are constrained by cost, size, and measurement speed, thereby rendering such laboratory implementations infeasible for real-world applications.
To overcome these shortcomings, one solution is to operate several radars at distinct subbands across a large bandwidth and fuse the radar data to improve the sensing resolution. 
However, practical multiband systems using commercially available mmWave radars face several implementation challenges and lack research on near-field SAR imaging.

Algorithms for radar imaging using multiband signal fusion have been investigated on numerous fronts over the last several decades \cite{cuomo1999ultrawide,tian2014sparse,zou2016matrix,wang2018wavenumber,zhang2014coherent,zhang2017multiple,tian2013multiband,li2008mft,sarkar1995mpa}. 
Autoregressive (AR) models have been employed for signal fusion \cite{cuomo1999ultrawide} using the root MUltiple SIgnal Classification (MUSIC) \cite{tian2014sparse} or the matrix-pencil approach \cite{zou2016matrix,wang2018wavenumber} to estimate the signal poles.
Recent approaches computed the signal poles of the AR model using Bayesian learning algorithms \cite{zhang2014coherent} and support vector regression \cite{zhang2017multiple}. 
Additionally, an all-phase fast Fourier transform (FFT) and iterative adaptive approach was proposed for signal-level fusion of de-chirped linear frequency modulated (LFM) signals \cite{tian2013multiband}.
Data-level fusion algorithms that require prefocusing prior to fusion have been explored for signal fusion in the frequency-wavenumber domain \cite{li2008mft,wang2018wavenumber}. 
In \cite{li2008mft}, a matrix Fourier transform (MFT) method was detailed for simple signal fusion in the $k$-domain; however, undesirable sidelobes in the range domain or $R$-domain were observed. 
To address this limitation, \cite{wang2018wavenumber} proposed a $k$-domain fusion algorithm using an iterative matrix-pencil algorithm (MPA) \cite{zou2016matrix,sarkar1995mpa} for improved microwave imaging. 
However, the MPA assumes a simplistic target model, and performance is degraded for complicated or intricate targets common in applications such as automatic target recognition (ATR) and concealed item detection.  
Alternatively, sparse representation algorithms are not suitable for most multiband scenarios because they require random, non-uniform sampling. 
Whereas sparse stepped frequency radar adheres to this constraint by sampling at random frequencies over a specified range \cite{zhang2011high}, multiband imaging scenarios typically do not have adequate subbands for sufficiently robust reconstruction with conventional sparse recovery algorithms \cite{zhang2014coherent}. 

Data-level multiband signal fusion requires phase coherence among the subbands, which is an issue for practical implementation. 
To ensure that the subbands are mutually coherent, \cite{tian2013multiband} details an algorithm using the all-phase FFT to estimate the incoherent phase (ICP) and compensate the ICP among subbands. 
In \cite{tian2014sparse}, a root-MUSIC method was proposed to estimate the ICP; however, the phase incurred by the frequency gap is not considered, and the algorithm demonstrated weak performance under noisy conditions. 
Later, an MPA-based approach was developed for ICP estimation and compensation, which yielded superior performance in low signal-to-noise ratio (SNR) scenarios \cite{zou2016matrix,wang2018wavenumber}. 

Data-driven algorithms that have dominated computer vision \cite{liu2022convnext}, specifically convolutional neural network (CNN) architectures, are gaining increasing interest for their applications in mmWave and THz imaging.
Several approaches employ CNN architectures in the image domain to achieve near-field SAR super-resolution for various tasks \cite{dai2021imaging,jing2022enhanced,smith2021An,vasileiou2022efficient,zhang2019target}.
A real-valued convolutional neural network (RV-CNN) was developed for near-field imaging to improve hand tracking resolution in \cite{smith2021An}.
Zhang \textit{et al.} proposed an encoder-decoder generative adversarial network (GAN) for through-the-wall radar imaging \cite{zhang2019target}.
A CNN was employed in \cite{dai2021imaging} to mitigate multiple-input multiple-output (MIMO) artifacts for image enhancement. 
A technique for mitigating artifacts caused by irregular SAR scanning patterns was developed using a CNN-GAN framework \cite{vasileiou2022efficient}. 
The first application of a complex-valued CNN (CV-CNN) was far-field Polarimetric SAR (PolSAR) image classification \cite{haensch2010complex}. 
Later, \cite{zhang2017complex} demonstrated the superiority of CV-CNNs over RV-CNN architectures for PolSAR processing at the image-level, and in \cite{jing2022enhanced}, a CV-CNN was proposed to operate on mmWave images, yielding improved image quality and sensing resolution. 
CNN architectures have also been used in conjunction with compressed sensing (CS) techniques for near-field mmWave imaging under sparse sampling \cite{wang2021tpssiNet,wang2021rmistnet}. 
At the signal level, data-driven algorithms have been employed for line spectra super-resolution and have been applied to far-field radar imaging \cite{izacard2021datadriven,pan2021complexFrequencyEstimation}.
Such techniques are advantageous over traditional line spectra super-resolution techniques, such as MUSIC \cite{baral2019impact}, because they are highly parallelizable and invertible, a key characteristic for implementation in near-field imaging algorithms such as the range migration algorithm (RMA) \cite{yanik2020development}. 
Although multiband signal fusion is a problem similar to line spectra super-resolution, deep learning-based fusion techniques have not been explored in previous studies. 

In this paper, we introduce a novel multiband signal fusion algorithm, $kR$-Net, using a hybrid, dual-domain CV-CNN approach for improved \mbox{3-D} SAR imaging. 
Under the Born approximation of the scattered field, we model the wavenumber domain signals as a sum of damped/undamped complex exponential functions sampled at each subband. 
Hence, the spectral representation of the multiband signal comprises the sum of the subband spectra.  
Recovering the missing data in the frequency gap between the subbands is a dual problem of wavenumber spectral enhancement. 
Applying a traditional convolutional architecture to the multiband signal in the wavenumber domain is challenging due to slow growth of the effective receptive field relative to the number of samples between the subbands, as described in Section \ref{subsubsec:dri_ablation_study}. 
As a result, many layers of the network lack the global context necessary to recover the equivalent wide bandwidth \cite{luo2016understanding}. 
However, forsaking processing in the wavenumber domain ignores the integral relationships between the known subbands and predicted samples in the frequency gap. 






To overcome these limitations, we introduce a novel CV-CNN architecture operating in both the wavenumber and wavenumber spectral domain, which outperforms equivalent approaches that act exclusively in one domain. 
Recently, the Fourier transform has been employed in deep learning to improve the efficiency of transformer architectures \cite{lee2021fnet} and to implement convolution in the Fourier domain, achieving global receptive fields \cite{chi2020fast}.
In \cite{souza2019hybrid}, a hybrid frequency/image domain network was developed for magnetic resonance (MR) image reconstruction using a \mbox{2-D} Fast Fourier Transform (FFT) to alternate between the image and frequency domains of the MR image.
Similarly, $kR$-Net applies a hybrid technique that interleaves forward and inverse FFTs between the convolutional layers in the CNN architecture. 
This allows the model to learn features in both the wavenumber domain, or $k$-domain, and its spectral domain, the $R$-domain.
This study is the first to leverage the relationships in both domains of multisinusoidal signals for deep learning-based signal processing.
Using this hybrid technique, the proposed CV-CNN architecture fuses multiband data in the wavenumber domain to form an equivalent wide bandwidth signal.
The model architecture also features a simplified residual block \cite{lim2017enhanced} and an improved complex-valued parametric activation function \cite{jing2022enhanced}. 
After the data from each subband are fused, the \mbox{3-D} SAR image can be recovered using an efficient Fourier-based algorithm \cite{yanik2020development}. 

The main technical contributions of this paper are as follows:
\begin{itemize}
\item[1)] A novel dual-domain CV-CNN architecture for multiband signal fusion in the $k$-domain and spectral super-resolution in the $R$-domain with improved convergence and performance compared with a network operating exclusively in either domain. 
\item[2)] Unlike previous multiband fusion techniques, $kR$-Net does not assume a small number of reflectors and boasts superior robustness for super-resolution of intricate, extended targets.
\item[3)] A multiband imaging prototype using two commercially available radars each with a bandwidth of 4 GHz operating at 60 GHz and 77 GHz and a synchronization technique similar to that in \cite{yanik2020development} for real-time multi-radar synchronization. 
\item[4)] Extensive experimental results, including simulation studies and validation using the proposed dual-radar prototype, verify that the proposed algorithm is capable of high-fidelity multiband signal fusion, outperforming existing techniques \cite{wang2018wavenumber,li2008mft}, while achieving high efficiency. Furthermore, we demonstrate the generalizability of our proposed technique to various frequency ranges and subband configurations in both the mmWave and THz spectra. 
\end{itemize}

The remainder of this article is organized as follows.
In Section \ref{sec:dri_signal_model}, the signal model for multiband radar signaling in the wavenumber domain, or $k$-domain, and wavenumber spectral domain, or $R$-domain, is formulated. 
In Section \ref{sec:dri_methods}, the proposed $kR$-Net is introduced, and the relevant details are discussed.
The development of a multiband imaging system is overviewed in Section \ref{sec:dri_system}, which leverages a multi-device synchronization approach for real-time measurements and high-precision spatial sampling. 
Simulation and empirical experiments are conducted, and the results are analyzed in Section \ref{sec:dri_results}, which is followed by the conclusions.


\section{Multiband Signal Model}
\label{sec:dri_signal_model}
\begin{figure}[ht]
    \centering
    \includegraphics[width=0.65\textwidth]{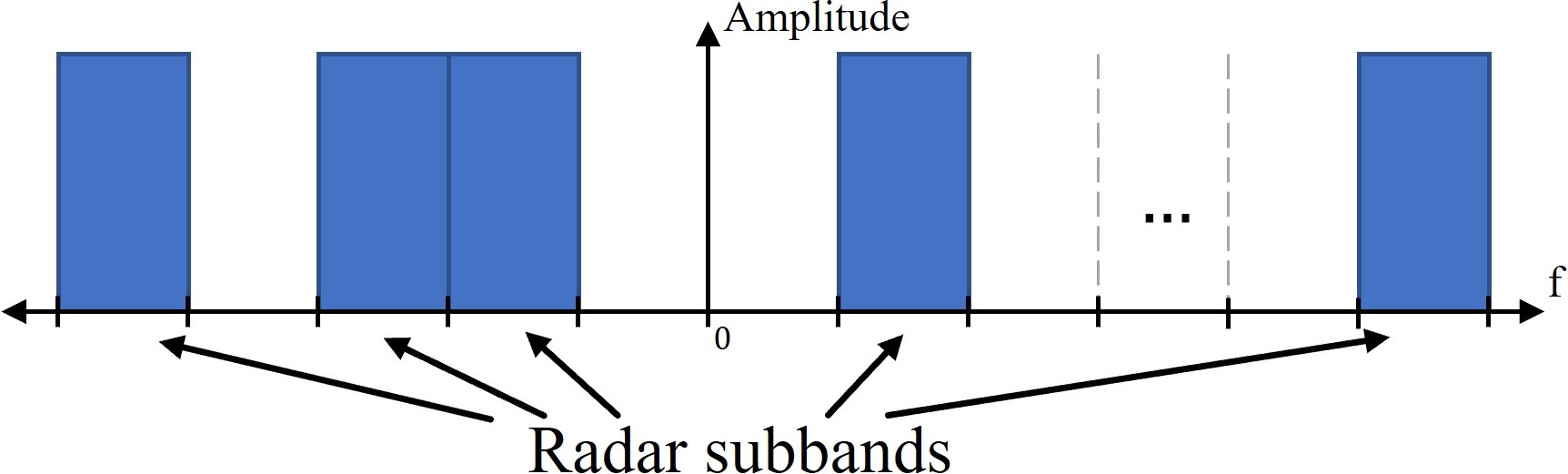}
    \caption{A baseband representation of multiband radar signaling where samples are gathered at multiple subbands separated by frequency gaps. The goal of multiband signal fusion is to recover the equivalent wideband signal spanning the entirety of the subbands.}
    \label{fig:general_multiband} 
\end{figure}

In this section, we formulate a signal model for multiband radar signaling. 
In the multiband sensing scenario, samples are taken across multiple subbands separated by frequency gaps, as shown in Fig. \ref{fig:general_multiband}, where the radar subbands represent the operating frequency ranges of the radars. 
It is important to note that the subsequent analysis and proposed algorithm assume a weak or constant relationship between the scattering properties and frequency across the entire bandwidth spanned by the subbands. 
To achieve the desired resolution, multiband signal fusion methods are applied to recover the unoccupied frequency bins and obtain the equivalent wideband signal spanning the entirety of the subbands. 

For the remainder of this study, we focus on the dual subband scenario emphasized in the existing literature \cite{cuomo1999ultrawide,tian2014sparse,zou2016matrix,wang2018wavenumber,zhang2014coherent,zhang2017multiple,tian2013multiband,li2008mft,sarkar1995mpa}.
However, the following analysis and proposed solution apply to the generalized multiband fusion problem portrayed in Fig. \ref{fig:general_multiband}, as detailed in Section \ref{subsubsec:dri_generalizability_study}. 
Additionally, although extrapolation of the signal beyond the highest and lowest subband frequencies has been proposed for previous techniques \cite{zhang2014coherent,zhang2017multiple}, we focus on developing an algorithm to reliably impute the missing signal in the frequency gap. 

In a generic near-field SAR or inverse-SAR (ISAR) scenario, two radars are mounted on a platform that scans a target scene for high-resolution imaging. 
Hence, provided proper system design, the synthetic aperture elements of both radars can overlap to produce a virtual monostatic element operating in the frequency ranges of both radars. 
Alternatively, the following fusion signal model can be achieved through other means, such as a monostatic wideband system sampled at several subbands to reduce sampling bandwidth, a MIMO dual-band system whose virtual elements of each subband overlap, or modeling the problem as fusion in the angular spatial wavenumber domain \cite{wang2018wavenumber}, provided the radars are sufficiently close to each other during scanning. 

Suppose the subbands start at frequencies $f_1$ and $f_2$, respectively, and illuminate a target in $x$-$y$-$z$ Cartesian space, where $z$ represents the downrange or range direction, and $x$-$y$ are known as the cross-range directions. 
Consider the monostatic element operating at both subbands and located at $(x',y',z')$ illuminating $N_t$ targets modeled as point scatterers, where the $i$-th target is located at $(x_i,y_i,z_i)$ with reflectivity $\alpha_i$. 
The wavenumber domain, or $k$-domain, response to a chirp signal at the first and second radars can be written as
\begin{align}
    \label{eq:radar1_response}
    s_1(n) &= \sum_{i=0}^{N_t-1} \alpha_i e^{-j2(k_1 + \Delta_k n)R_i}, \ n = 0, \dots, N_k-1, \\
    s_2(n) &= \sum_{i=0}^{N_t-1} \alpha_i e^{-j2(k_2 + \Delta_k n)R_i}, \ n = 0, \dots, N_k-1, 
    \label{eq:radar2_response}
\end{align}
where $k_1$ and $k_2$ are the wavenumbers corresponding to the starting frequencies $f_1$ and $f_2$, respectively, $k = 2\pi f/c$, $\Delta_k$ is the wavenumber sampling interval, $n$ is the time sample index, $N_k$ is the number of samples in each subband, and $R_i$ is the distance from the radar to the $i$-th scatterer, which is expressed as 
\begin{equation}
    \label{eq:R_i}
    R_i = \left[ (x' - x_i)^2 + (y' - y_i)^2 + (z' - z_i)^2 \right]^\frac{1}{2}.
\end{equation}
Although the sampling conditions, sampling rate $\Delta_k$ and number of samples $N_k$, are considered identical across subbands for simplicity, this is not a strictly necessary condition as the subbands could have different sampling conditions. 
While different values of $N_k$ raises a trivial issue, different sample rates among subbands will need to be compensated such that the spectral domains are coincident. 
In addition, the proposed signal model assumes that the scattering parameters, $\alpha_i$, are frequency-independent. 
However, in a real scenario, scattering properties of the various materials in the target scene vary across subbands, a phenomenon that may not be adequately modeled by (\ref{eq:radar1_response}) and (\ref{eq:radar2_response2}), depending on the material properties and frequency ranges of the subbands. 
Since the proposed training data scheme assumes the frequency-independence of $\alpha_i$, networks trained on these data are limited because of this assumption, as discussed in Section \ref{subsec:dri_training_details}. 

Both $s_1(\cdot)$ and $s_2(\cdot)$ are considered multisinusoidal signals because they are composed of a superposition of scaled complex exponential functions, whose frequencies are determined by the ranges $R_i$. 
Hence, the wavenumber spectral domain, known as the range domain or $R$-domain, exhibits peaks at positions corresponding to the ranges $R_i$.
Defining $\Delta_B \triangleq k_2 - k_1$, the difference between the starting wavenumbers, the signal at the second subband can be rewritten with respect to $k_1$ and different indexing as
\begin{equation}
    \label{eq:radar2_response2}
    s_2(n') = \sum_{i=0}^{N_t-1} \alpha_i e^{-j2(k_1 + \Delta_k n')R_i}, \ n' = \Tilde{N}, \dots, N,
\end{equation}
such that $\Tilde{N} \triangleq \Delta_B/\Delta_k$ is the offset between subbands 1 and 2, where $N \triangleq \Tilde{N} + N_k - 1$, $n' = n + \Tilde{N}$, and $\Tilde{N} > N_k$.
We assume that $\Tilde{N}$ is an integer based on the choices of $\Delta_k$ and $\Delta_B$, although the derivation is valid regardless.

\begin{figure}[th]
    \centering
    \includegraphics[width=0.75\textwidth]{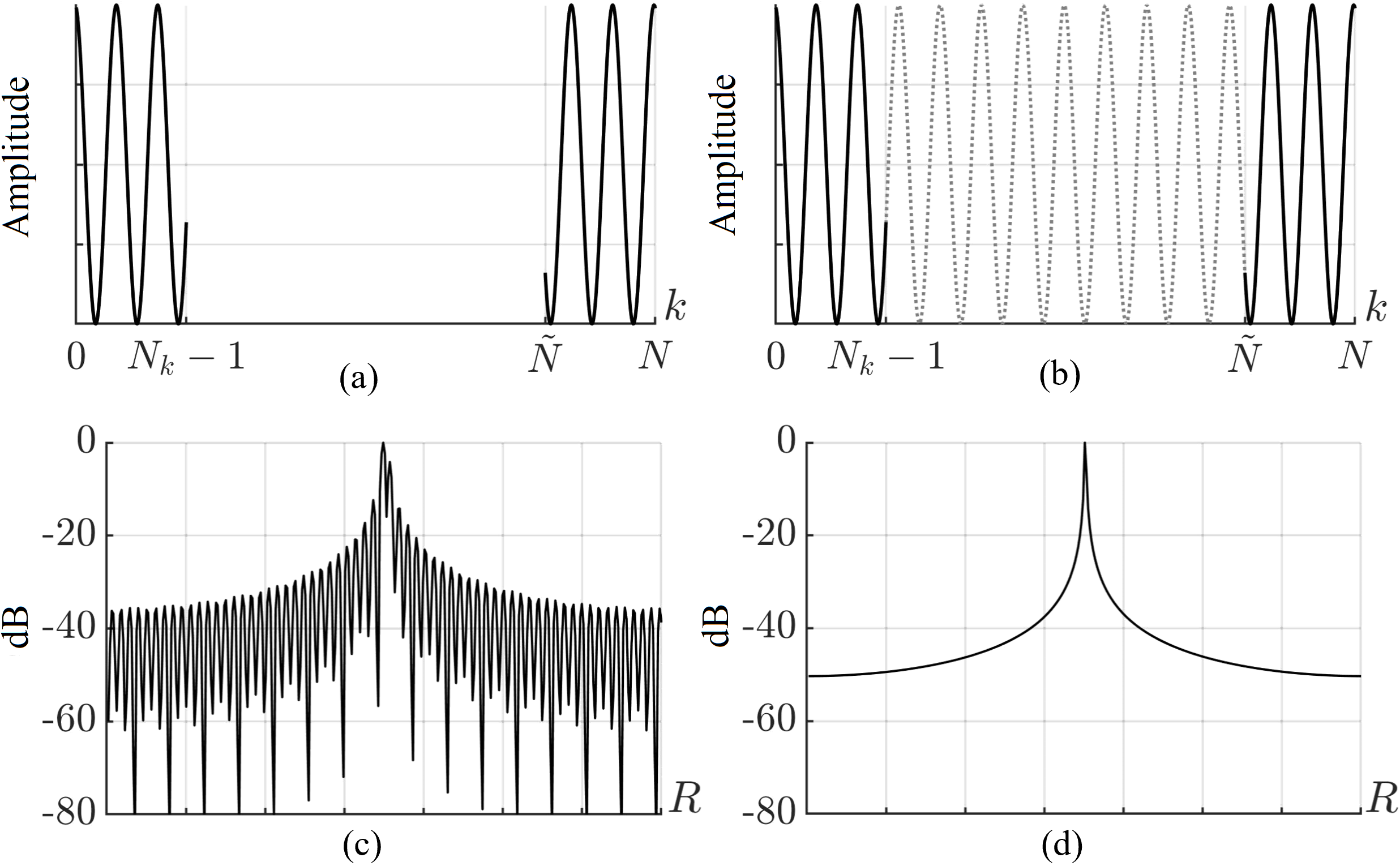}
    \caption{Multiband scenario with two subbands in the $k$-domain and $R$-domain. (a) $k$-domain non-contiguous dual-band signal, (b) $k$-domain ideal full-band signal, (c) $R$-domain spectrum of the non-contiguous dual-band signal with zero-padding, (d) $R$-domain spectrum of the ideal full-band signal.}
    \label{fig:multiband_scenario} 
\end{figure}

From (\ref{eq:radar1_response}) and (\ref{eq:radar2_response2}), the multiband scenario can be understood as a multisinusoidal signal sampled across several disjoint regions offset by $\Tilde{N}$.
We refer to the two subband scenario as \textit{non-contiguous dual-band}, as shown in Fig. \ref{fig:multiband_scenario}a.
As the two subbands are not coherent in a practical implementation, we implement the algorithm developed in \cite{wang2018wavenumber} to efficiently estimate the ICP and compensate each subband accordingly. 
Additional details on the mutual coherency among subbands can be found in \cite{tian2013multiband,tian2014sparse,zou2016matrix,wang2018wavenumber}. 
The signal in the $k$-domain can be represented in the $R$-domain, or wavenumber spectral domain, by taking the Fourier transform. 
Limiting the sampling in each subband to $N_k$ corresponds to a convolution in the $R$-domain with a discrete sinc or Dirichlet kernel of width $1/N_k$, resulting in smearing of the spectral information and causing closely spaced peaks to bend together \cite{izacard2021datadriven}.
Given the structure of the multiband signal, the $R$-domain spectrum is the sum of the spectra for each subband if the frequency gap is ignored. 
Because of the sinc-effect and phase shift in the $R$-domain corresponding to the $\Tilde{N}$ sample shift in the $k$-domain for each of the $N_t$ reflectors, the non-contiguous dual-band signal in the $R$-domain suffers from artifacts/sidelobes as the frequency gap between the subbands is neglected \cite{wang2018wavenumber}, as shown in Fig. \ref{fig:multiband_scenario}c.
This analysis is identical to the MFT \cite{li2008mft}, which results in images degraded by increased sidelobes in the range direction. 
Comparatively, the $R$-domain spectrum of the ideal \textit{full-band} signal (Fig. \ref{fig:multiband_scenario}d) does not contain spurious peaks that would distort the images recovered from the fused signal, thereby achieving an improved resolution compared with each subband and the MFT approach.

\subsection{Existing Methods for Multiband Signal Fusion}
\label{subsec:dri_existing_methods}
The objective of multiband signal fusion is to impute the bandwidth between the subbands (interpolating between the subbands or extrapolating the missing samples) in the $k$-domain to acquire the ideal full-band signal shown in Fig. \ref{fig:multiband_scenario}b, where the dotted portion represents the signal in the frequency gap. 
Methods for recovering the missing wavenumber domain data from $N_k$ to $\Tilde{N}-1$ apply MUSIC \cite{tian2014sparse} or MPA \cite{zou2016matrix,wang2018wavenumber} to estimate the signal poles in an all-pole model. 
However, these approaches assume that the estimated model order of $s_1(\cdot)$ and $s_2(\cdot)$, $\hat{N}_t \approx N_t$, is small compared to $N_k$.
From the analysis in \cite{sarkar1995mpa}, $\hat{N}_t$, the estimated number of targets in the scene, must be chosen such that $\hat{N}_t < \text{round}(N_k/3)$ for the MPA \cite{wang2018wavenumber}.
After the $\hat{N}_t$ signal poles and coefficients are computed, the missing samples can be estimated. 
However, high-resolution near-field SAR often requires imaging of intricate, continuous objects modeled by thousands or millions of point scatterers, or $N_t \gg N_k$ \cite{smith2020nearfieldisar,batra2021short}. 
As a result, traditional approaches such as the MPA assume simplistic targets, thereby neglecting high-frequency features of the target, and are unable to faithfully recover the multiband signals. 

For clarity in the remainder of this paper, we propose new terminology to describe the spectral composition of radar target scenes based on the portion of the baseband bandwidth occupied by the reflected signal. 
A target consisting of fewer reflectors than the number of frequency samples $(N_t < N_k)$ over the specified frequency range has a \textit{low-bandwidth} relative to the bandwidth of the system. 
Low-bandwidth targets have a low model order and a low-rank sample covariance matrix, allowing conventional algorithms to adequately approximate signal poles and coefficients. 
In contrast, a target consisting of a large number of reflectors relative to the frequency sampling $(N_t \gg N_k)$ is a \textit{high-bandwidth} target, which is typical in most security and industrial applications of near-field SAR imaging. 
High-bandwidth targets contain intricate, high-frequency spatial features and have not been addressed for multiband signal fusion in previous studies. 

\begin{figure*}[t]
    \centering
    \includegraphics[width=\textwidth]{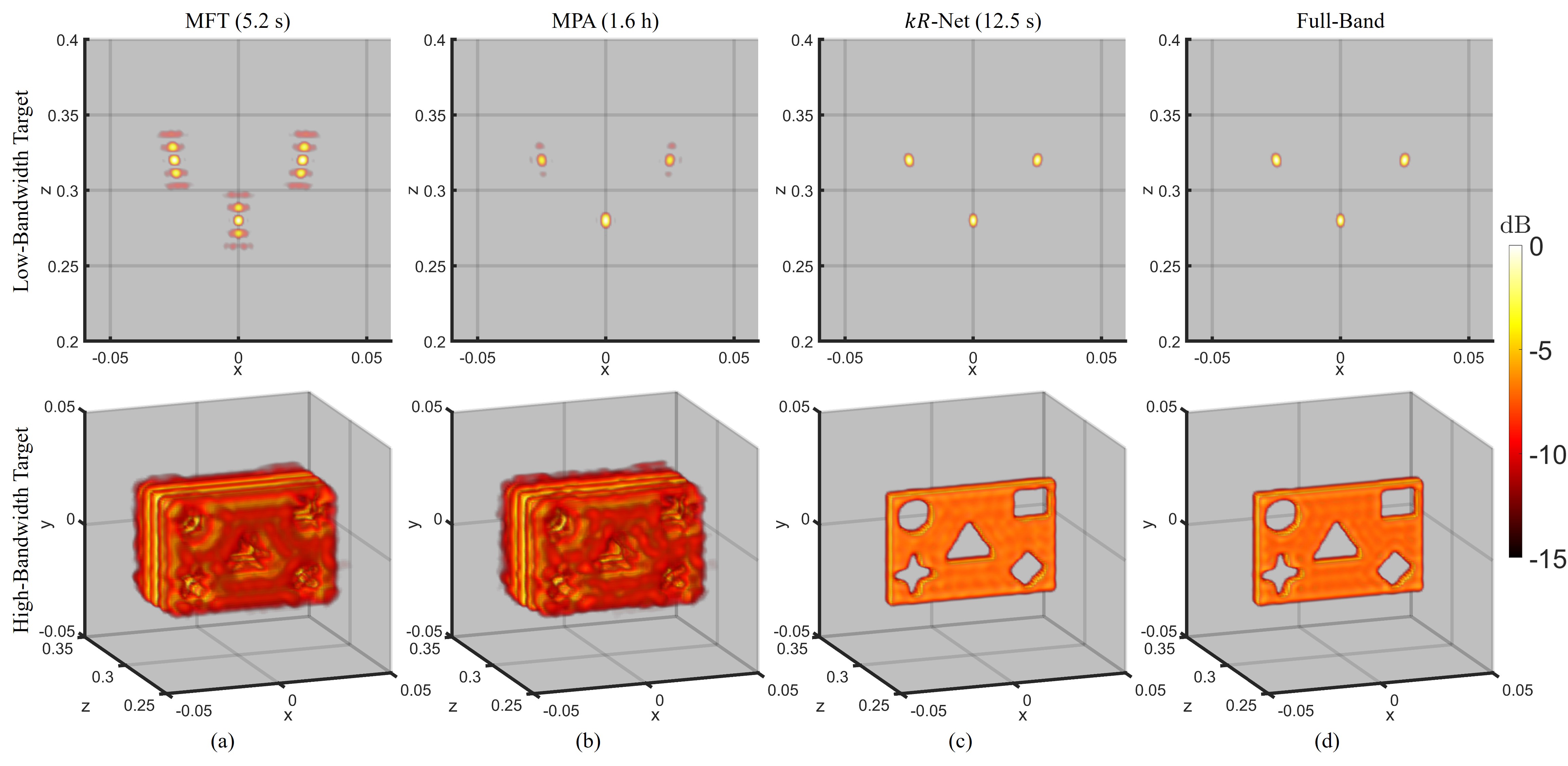}
    \caption{Demonstration of the limitations of the matrix Fourier transform (MFT) and matrix-pencil algorithm (MPA) for multiband signal fusion using two targets in simulation: (top) three point targets and (bottom) rectangle with various cutout shapes. (a) The MFT, which requires 5.2 s, suffers from significant sidelobes since it does not compensate for the missing samples in the frequency gap. (b) Although the MPA, which requires 1.6 h for a GPU implementation, achieves adequate reconstruction for the low-bandwidth target (top), its performance degrades for a high-bandwidth target (bottom). (c) The proposed $kR$-Net, which requires 12.5 s, recovers an image that closely resembles the (d) ideal reference image corresponding the full-band scenario as in Figs. \ref{fig:multiband_scenario}b and \ref{fig:multiband_scenario}d.}
    \label{fig:ex_mft_mpa_ours} 
\end{figure*}

To illustrate this phenomenon, we consider two cases: (1) a low-bandwidth target consisting of 3 point reflectors and (2) a \mbox{3-D} model of a rectangle with various cutout shapes to constitute a high-bandwidth target, as shown in the first and second rows of Fig. \ref{fig:ex_mft_mpa_ours}, respectively. 
All images are reconstructed using the RMA \cite{yanik2020development,sheen2001three,smith2022efficient,vu2022fourier} with the SAR scanning parameters detailed in Section \ref{subsubsec:dri_sim2_cutout1}. 
The MFT \cite{li2008mft} is applied to fuse the multiband data, yielding the images in the left column (Fig. \ref{fig:ex_mft_mpa_ours}a). 
As expected, since the MFT does not account for the frequency gap between the subbands, increased sidelobes are observed along the $z$-direction. 
However, the quality of images recovered using the MPA \cite{wang2018wavenumber} varies significantly. 
For the simple low-bandwidth target in the top row $(N_t = 3, N_k = 64)$, the MPA recovers each point with minimal undesirable sidelobes compared to the MFT. 
However, although its performance is better than the MFT, the MPA is plagued by considerable degradation for the high-bandwidth target, where $N_t$ is on the order of thousands and $N_t \gg N_k$, as the high-bandwidth features of the target are not adequately modeled by the MPA. 
For an equitable comparison, a parallelized GPU implementation of the MPA is employed \cite{zou2016matrix,wang2018wavenumber}. 
Hence, the computation time required for the GPU-implemented MPA is still 1.6 h for this example, deeming it unfit for many applications demanding rapid imaging, such as packing and security screening.
By comparison, the MFT and $kR$-Net boast computation times of 5.2 s and 12.5 s, respectively, enabling many common mmWave imaging solutions. 
In the right column (Fig. \ref{fig:ex_mft_mpa_ours}c), the imaging results obtained using the proposed $kR$-Net demonstrate robustness for both targets by achieving focusing performance comparable to the ideal, full-band scenario. 
Notably, the intricate features of the high-bandwidth target are retained, and the resolution in the $z$-direction is significantly improved compared with the MFT and MPA. 

Multiband signal fusion can be posed as a spectral super-resolution/restoration problem in the $R$-domain, the dual to imputation in the $k$-domain.
As shown in Figs. \ref{fig:multiband_scenario}c and \ref{fig:multiband_scenario}d, $R$-domain super-resolution of the $N_t$ peaks corresponds to imputation of the full bandwidth \cite{wang2019multi}. 
Deep learning-based solutions have proven successful in similar spectral-enhancement problems on radar images \cite{dai2021imaging,jing2022enhanced,smith2021An,vasileiou2022efficient,zhang2019target} and multisinusoidal line spectra \cite{izacard2021datadriven,pan2021complexFrequencyEstimation}, achieving resolutions exceeding the theoretical limitations. 
However, data-driven approaches have not been applied to multiband signals to achieve joint $k$-domain imputation and $R$-domain super-resolution. 
Because the multiband fusion problem has distinct features in the $k$-domain and $R$-domain, we propose a hybrid approach that operates in both domains. 


\section{Proposed Architecture of $kR$-Net for Improved 3-D Multiband SAR Imaging}
\label{sec:dri_methods}

In this section, we introduce a novel dual-domain CV-CNN architecture, referred to as $kR$-Net, to perform efficient multiband fusion for improved \mbox{3-D} near-field SAR imaging. 
The proposed framework alternates between operating in the $k$-domain and $R$-domain, allowing the network to learn the unique characteristics inherent to each domain. 
Compared with learning in only one domain, $kR$-Net demonstrates superior convergence and quantitative performance, as discussed in Section \ref{sec:dri_results}.
Additionally, the proposed algorithm is robust for low- and high-bandwidth imaging scenarios, which are common in many realistic applications.

\begin{figure*}[t]
    \centering
    \includegraphics[width=\textwidth]{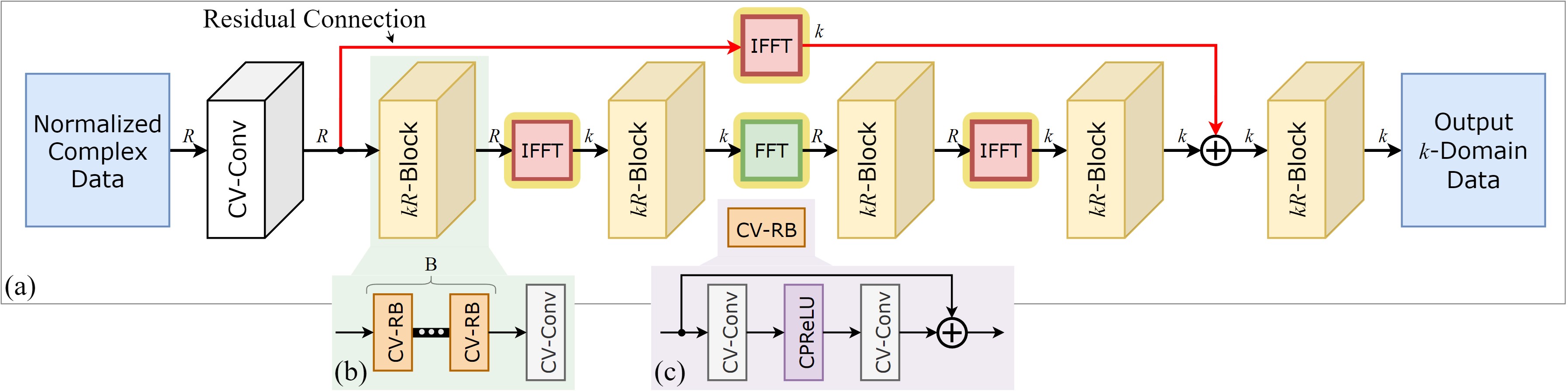}
    \caption{Architecture of the (a) hybrid, dual-domain $kR$-Net. Proposed addition of the domain transformation blocks (highlighted: FFT and IFFT) allows the network to learn important features of the signal in both the $k$- and $R$-domains improving on conventional CNN approaches. The novel, hybrid approach using FFT and IFFT blocks throughout the network, achieves superior multiband signal fusion and spectral super-resolution compared with a model without the Fourier operations. Architecture of the (b) $kR$-Block and (c) CV-RB. The $kR$-block consists of $B$ CV-RBs in cascade followed by a CV-Conv layer. Each CV-RB is a residual convolution block with a single CPReLU activation.}
    \label{fig:kRNet_overview} 
\end{figure*}

\subsection{Framework of $kR$-Net}
The architecture of $kR$-Net is shown in Fig. \ref{fig:kRNet_overview}a, where the signal domain is denoted at each connection as $k$ or $R$ for the wavenumber domain or wavenumber spectral domain, respectively, and the forward and inverse fast Fourier transform are denoted by FFT and IFFT, respectively. 
The input is given as the $R$-domain spectrum of the wavenumber domain samples and is processed in both domains by residual $kR$-blocks. 
After the residual connection, an additional $kR$-block and complex-valued convolution (CV-Conv) layer processes the signal before it is output in the $k$-domain.
The specific designs of each module are detailed as follows.

\subsubsection{Input Layer and Residual Connection}
\label{subsubsec:input_layer}
Rather than layering the real and complex parts of the signal \cite{smith2021sterile,izacard2021datadriven} or employing a two-path network \cite{wang2021tpssiNet}, the inputs to $kR$-Net are complex-valued signals of length $N$. 
The complex-valued input signals are normalized in the $R$-domain by the magnitude min-max norm before being passed to the first CV-Conv layer as
\begin{equation}
    \label{eq:min_max_norm}
    \mathbf{z} = \frac{\mathbf{x} - ||\mathbf{x}||_\text{min}}{||\mathbf{x}||_\text{max} - ||\mathbf{x}||_\text{min}},
\end{equation}
where $||\mathbf{x}||_\text{min}$ and $||\mathbf{x}||_\text{max}$ are the minimum and maximum values of the magnitude of $\mathbf{x}$. 
Hence, the phase of $\mathbf{x}$ remains unmodified, while the magnitude is scaled to be consistently between 0 and 1. 

Since the network expects an input in the $R$-domain, the FFT of the $k$-domain data is computed prior to the input to the network with zero-padding between the subbands.
Complex-valued convolution extends the convolution operation employed by CNNs to complex input data. 
To implement a CV-Conv layer, the convolution kernel matrix must be complex-valued. 
However, because the convolution between two complex-valued tensors is generally unsupported by deep learning software, we decompose the input signal $\mathbf{x} = \mathbf{x}_R + j\mathbf{x}_I$ into real and imaginary parts.
Similarly, by expressing the kernel as $\mathbf{M} = \mathbf{M}_R + j\mathbf{M}_I$, the complex-valued convolution can be written, neglecting the bias terms, as
\begin{equation}
    \label{eq:cv_conv}
    \mathbf{x} \circledast \mathbf{M} = \mathbf{x}_R \circledast \mathbf{M}_R - \mathbf{x}_I \circledast \mathbf{M}_I + j(\mathbf{x}_R \circledast \mathbf{M}_I + \mathbf{x}_I \circledast \mathbf{M}_R),
\end{equation}
where $\circledast$ denotes the real-valued convolution operation. 
By decomposing the convolution in this manner, complex-valued convolution can be computed using existing techniques operating on the real and imaginary parts of the input signal and kernel. 
The real and imaginary parts of the weight matrix $\mathbf{M}$ can be implemented as real-valued matrices according to (\ref{eq:cv_conv}), and their values are determined by complex-valued backpropagation following the convention for CV-CNNs \cite{zhang2017complex,jing2022enhanced}. 

The CV-Conv layer is a general-purpose complex-valued convolution layer defined with a kernel size $K$ and zero-padding such that the signal length of $N$ is preserved at the output, $C_\text{in}$ input channels, and $C_\text{out}$ output channels.
The first layer of $kR$-Net is a CV-Conv layer with 1 input channel and $F$ output channels, where $F$ is the number of feature channels and is constant throughout the network.
After the first CV-Conv layer, the intermediate representations are fed through an IFFT block in the residual pass-forward connection, as shown in red in Fig. \ref{fig:kRNet_overview}a. 
The residual connection preserves the information at the known subbands, and the network demonstrates superior empirical performance with the proposed configuration than without the pass-forward connection. 

\subsubsection{$kR$-Blocks and Domain Transformation Blocks}
\label{subsubsec:kR_block}
The $kR$-Block is composed of a cascade of complex-valued residual blocks (CV-RBs) followed by a single CV-Conv layer, as shown in Fig. \ref{fig:kRNet_overview}b. 
Each $kR$-Block operates on the signal in either the $k$-domain or $R$-domain, as the signal alternates between the two domains throughout the $kR$-Net. 
Furthermore, because the domain transformation blocks (FFT and IFFT) are fully differentiable, they can be treated as conventional layers in the network, and gradient backpropagation can be easily implemented \cite{jing2022enhanced}.
The Fourier operations are performed across each activation map and normalized to make the FFT and IFFT orthonormal.
Based on the convolution properties of the Fourier transform, convolution in one domain can be viewed as multiplication in the other domain. 
In this sense, applying a CV-Conv layer in the $k$-domain can be considered a fully connected layer in the $R$-domain. 
However, as illustrated later in Section \ref{subsubsec:dri_ablation_study}, our hybrid, dual-domain approach outperforms a network operating exclusively in the $k$-domain or $R$-domain in terms of convergence and numerical performance. 
Hence, compared with conventional CNN models, the addition of domain transformation blocks throughout the network is key to improving multiband fusion performance. 

The CV-RB architecture is shown in Fig. \ref{fig:kRNet_overview}c. 
Inspired by \cite{lim2017enhanced}, the residual block consists of two convolution layers separated by an activation function: the complex parametric rectified linear unit (CPReLU) \cite{jing2022enhanced}. 
Compared with the original ResNet residual block \cite{he2016deep}, the proposed residual block removes batch normalization and empirically outperforms a bottleneck residual architecture \cite{liu2022convnext} for multiband signal fusion. 

The CPReLU activation function is selected over alternatives, such as the complex ReLU (CReLU) \cite{gao2018enhanced}, which computes the sum of the ReLU operation on the real and imaginary values separately, as the CPReLU has an improved activation over the complex domain. 
Using the notation employed in (\ref{eq:cv_conv}), the CPReLU can be expressed as 
\begin{align}
    \label{eq:CPReLU}
    \begin{split}
        \text{CPReLU}(\mathbf{x}) &= \max(0, \mathbf{x}_R) + \eta_R \min(0, \mathbf{x}_R) \\ &+ j\left( \max(0, \mathbf{x}_I) + \eta_I \min(0, \mathbf{x}_I) \right) ,
    \end{split}
\end{align}
where the parameters $\eta_R$ and $\eta_I$ are learned during the training stage of the network for each CPReLU instance \cite{jing2022enhanced}. 
Because $\eta_R$ and $\eta_I$ are learned independently, different layers of the network may learn different representations of the signal in amplitude and phase, aiding network robustness. 
The CPReLU can be understood as a complex domain parametric rectified linear unit (PReLU), which extends the traditional ReLU into the negative input domain to overcome gradient saturation for negative activation values. 
In the CPReLU, the real and imaginary parts of $\mathbf{x}$ are independently processed by a PReLU and the output is complex-valued. 
In the complex domain, this corresponds to retaining information in all four quadrants, corresponding to all combinations of positive and negative real and imaginary activation values, as detailed in \cite{jing2022enhanced}. 

After the first four $kR$-Blocks, the residual connection is made in the $k$-domain followed by another $kR$-Block and CV-Conv layer before being output, as shown in Fig. \ref{fig:kRNet_overview}a.
The number of $kR$-Blocks and domain transformations was investigated empirically, and the proposed configuration yielded the optimal numerical performance. 
However, further investigation of alternate architectures and deeper neural networks is a promising route for future research. 
In the spirit of \cite{liu2022convnext}, multiple values for the convolution kernel size $K$ were investigated, and the optimal value was determined empirically to be $K = 5$.
The number of feature maps throughout the network is chosen as $F = 32$, and the number of CV-RBs for each $kR$-block is set as $B = 8$.
$kR$-Net comprises 86 CV-Conv layers and 866324 learnable parameters. 
Multiband signal fusion is performed by $kR$-Net on a signal of length $N$, yielding a fused signal in the $k$-domain of equivalent size. 
As discussed in Section \ref{subsubsec:dri_generalizability_study}, a network is trained for a specific multiband signal fusion scenario and must be retrained for application to alternate subband configurations (placement and range of subbands, sampling conditions, length $N$ of full-band signal, etc.).


\subsection{Training Details}
\label{subsec:dri_training_details}
The weights of the network are calibrated using an Adam optimizer with a learning rate of $1 \times 10^{-4}$, $\beta_1 = 0.9$, and $\beta_2 = 0.999$.
Training is performed on a single RTX3090 GPU with 24 GB of memory with a batch size of 1024 and L1 loss criterion. 
The complex-valued loss term is defined as 
\begin{equation}
    \label{eq:l1_loss_cv}
    \mathcal{L} = \sum_{\ell=0}^{N-1} \biggr[ |\hat{s}(\ell)_R - s(\ell)_R| + |\hat{s}(\ell)_I - s(\ell)_I| \biggr],
\end{equation}
where $\hat{s}(\ell)$ are the predicted signals output from the network corresponding to the full-band ground-truth vectors $s(\ell)$ and the subscripts denote the real and imaginary parts of the signals. 
The complex components are processed separately, similar to the approaches in \cite{jing2022enhanced,zhang2017complex}, by traditional L1 distance metrics, and the real-valued result is the sum of the two L1 losses from the real and imaginary parts of the predicted signal with the ground-truth signal. 

More advanced loss functions were investigated, such as the L1/L2 difference between the ground-truth signal and intermediate representations throughout the network or loss between sample covariance matrices. 
These loss functions were tested in conjunction with alternate configurations, such as varying the number of $kR$-Blocks or removing the residual connection. 
However, the architecture detailed in Fig. \ref{fig:kRNet_overview}a demonstrated superior numerical performance in both training and testing with real multiband SAR data. 
Nevertheless, future investigations into neural network design based on statistical signal processing principles will likely facilitate additional insights and promising results for signal processing problems. 
Incorporating a hybrid, data-driven signal processing approach is a promising direction for similar future efforts. 

\subsubsection{Training and Testing Datasets}
\label{subsubsec:dri_dataset}
Since there is no publicly available dataset for near-field multiband SAR imaging, we generate training and testing datasets by simulating the response to a multiband LFM radar.
The ideal noiseless full-band signals spanning both subbands and the frequency gap in the $k$-domain are generated as 
\begin{equation}
    \label{eq:fb_sim}
    s(\ell) = \sum_{i=0}^{N_t - 1} \alpha_i e^{-j 2(k_1 + \Delta_k \ell) R_i}, \ell = 0, \dots, N - 1,
\end{equation}
where $\alpha_i$ values are selected from a complex normal distribution and $R_i$ values are chosen from a uniform distribution spanning the unambiguous range of the radar. 
After computing the full-band signals $s(\cdot)$ the multiband signals $\hat{s}(\cdot)$ are obtained by nullifying the samples in the frequency gap as
\begin{equation}
    \hat{s}(\ell)=\begin{cases}
          s(\ell) \quad & \, \ell \in [0, N_k - 1] \cup [\Tilde{N}, N - 1] \\
          0 \quad & \, \ell \in [N_k, \Tilde{N} - 1] \\
     \end{cases}.
\end{equation}
The multiband signals are then corrupted with complex additive Gaussian white noise (AWGN) in each subband. 
The noisy multiband signals are used as the input to $kR$-Net after taking the FFT and employing the normalization process detailed earlier.
Each noisy multiband signal is treated as a feature vector with a corresponding label vector consisting of the ideal full-band signal in (\ref{eq:fb_sim}). 

To train the network, 1048576 samples are independently generated with $N_t$ target reflectors, where $N_t$ is randomly selected between 1 and 200. 
The SNR for each sample is selected on a continuous uniform distribution from -10 to 30 dB. 
A validation set of 2048 samples is generated using the same procedure. 
Assuming a realistic scenario with two radars with starting frequencies $f_1 = 60$ GHz and $f_2 =$ 77 GHz, where each radar has a bandwidth of $B =$ 4 GHz, we set $N_k =$ 64 and $\Delta_f =$ 62.5 MHz. 
Hence, $\Tilde{N} = 272$ and $N = 336$, and the low-rank assumption of the MPA, $\hat{N}_t < \text{round}(N_k/3)$, will often be invalid if the target is high-bandwidth and consists of many reflectors. 
Although this study employs 60 GHz and 77 GHz radars, the proposed algorithm can easily be extended to other multiband configurations. 
For practical implementation, federal communications commission (FCC) licensing limits certain combinations of subbands based on application, but the algorithm and concepts derived in this study are applicable across various subband configurations. 
By training on this dataset, the proposed algorithm learns to perform multiband fusion for high-bandwidth targets. 

It is important to note that the proposed signal model and dataset generation scheme impose limitations on the model and its generalizability. 
In a practical implementation, imperfections that are not modeled in (\ref{eq:fb_sim}), such as frequency dependence of the scattering parameters, $\alpha_i$, device non-linearity, clutter, non-Gaussian noises, and different antenna distortions among subbands, may impact the reconstruction quality. 
Hence, networks trained on these data may display different robustness depending on the imaging scenario, placing additional importance on the system design and calibration. 
Although a model trained for a specific multiband scenario will not generalize well to other subband configurations, the proposed model architecture demonstrates the ability to generalize to various multiband imaging schemes when trained with appropriate datasets, as detailed in Section \ref{subsubsec:dri_generalizability_study}. 

An alternative dataset was considered consisting of extended, solid targets, such as those shown in Figs. \ref{fig:ex_mft_mpa_ours} and \ref{fig:dri_sim6}. 
However, because there is no sufficiently diverse dataset of such multiband SAR data, the network did not generalize well across different shapes. 
Additionally, a network was first trained on a dataset of randomly placed point reflectors, as in (\ref{eq:fb_sim}), and then fine-tuned on data from solid targets, but a performance improvement was not observed. 
Further development of diverse mmWave datasets will be essential to the advancement of joint signal processing and data-driven algorithms and is a promising future direction. 

\begin{figure}[h]
    \centering
    \includegraphics[width=0.55\textwidth]{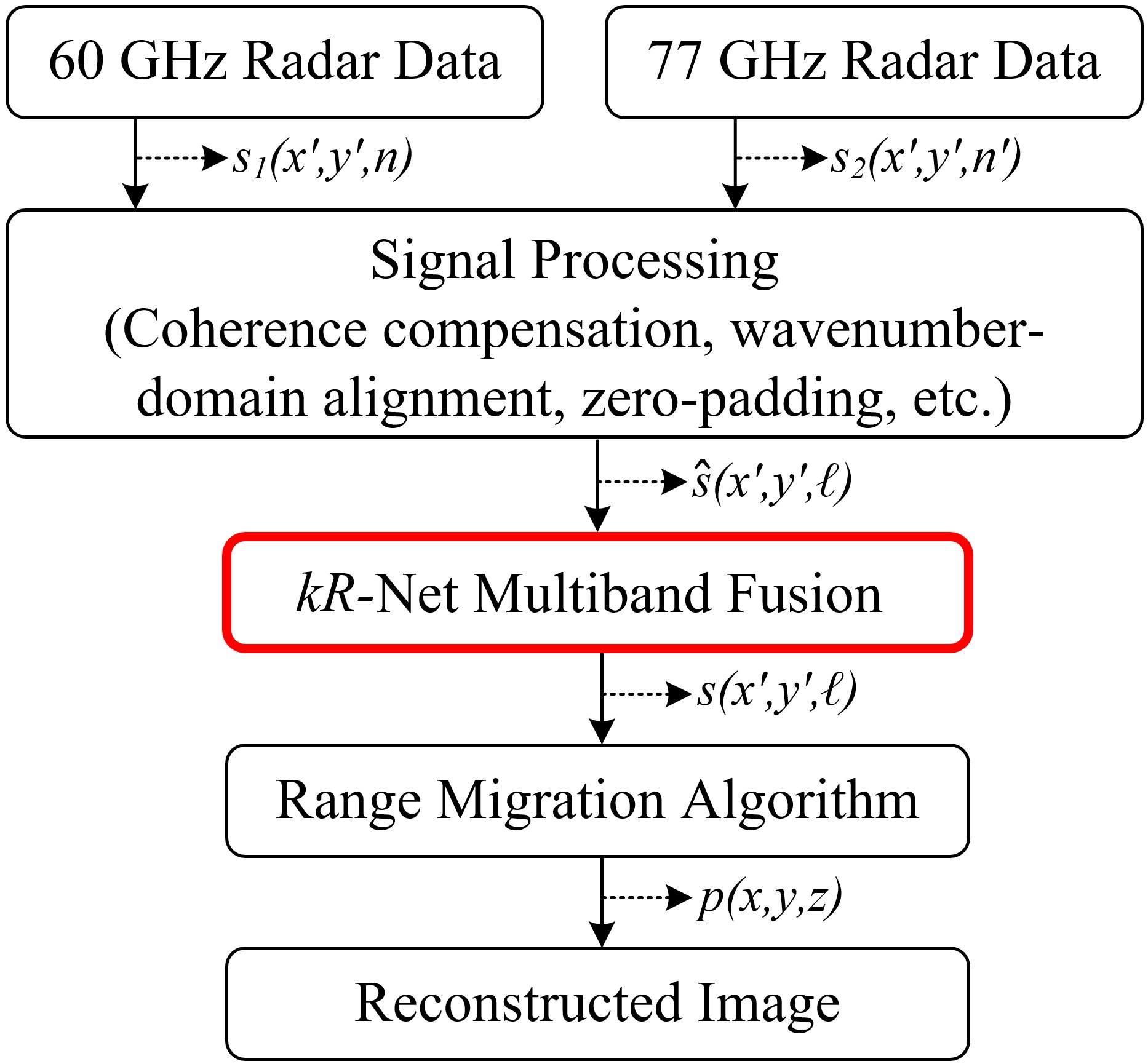}
    \caption{Multiband signal fusion pipeline for high-resolution \mbox{3-D} near-field imaging. The data from both subbands are fused using the proposed $kR$-Net producing a high-fidelity reconstruction for both low- and high-bandwidth targets.}
    \label{fig:dri_flow} 
\end{figure}

\subsection{Imaging Implementation}
\label{subsec:dri_rma_algo}
After multiband fusion is performed using $kR$-Net, the SAR image is reconstructed from the fused data. 
A summary of the imaging pipeline is presented in Fig. \ref{fig:dri_flow}. 
After the data are collected from both radars, preprocessing steps are necessary to ensure signal coherence and align the data in the $k$-domain. 
We implement the ICP compensation algorithm detailed in \cite{wang2018wavenumber} and set consistent sampling parameters across radars. 
The proposed algorithm is valid for both collocated and noncollocated antennas if the spatial wavenumber domains of all radars are sufficiently coincident, implying that both radars have similar illumination of the target. 
Applying $kR$-Net to the multiband signal is advantageous compared with classical signal processing algorithms as $kR$-Net is highly parallelizable and can efficiently perform signal fusion for many samples. 
The RMA is applied after the signal fusion step to produce a high-resolution \mbox{3-D} image \cite{yanik2020development,smith2022efficient,yanik2019cascaded}.
Compared to conventional signal processing-based algorithms for multiband fusion, the proposed $kR$-Net yields superior imaging performance and demonstrates robustness for the realistic case of high-bandwidth, intricate targets.

\section{Multiband Imaging System}
\label{sec:dri_system}
This section provides an overview of the implementation of the multiband imaging prototype using commercially available mmWave radars. 
Whereas prior research on near-field multiband radar imaging has employed sophisticated laboratory equipment \cite{wang2018wavenumber,tian2013multiband}, which is not suitable for many practical applications, we introduce a highly-integrated system that employs commercially available equipment for multiband near-field SAR. 
The proposed testbed uses two mmWave radars operating at distinct subbands and introduces a synchronization strategy to achieve efficient data collection.

\begin{figure}[h]
    \centering
    \includegraphics[width=0.75\textwidth]{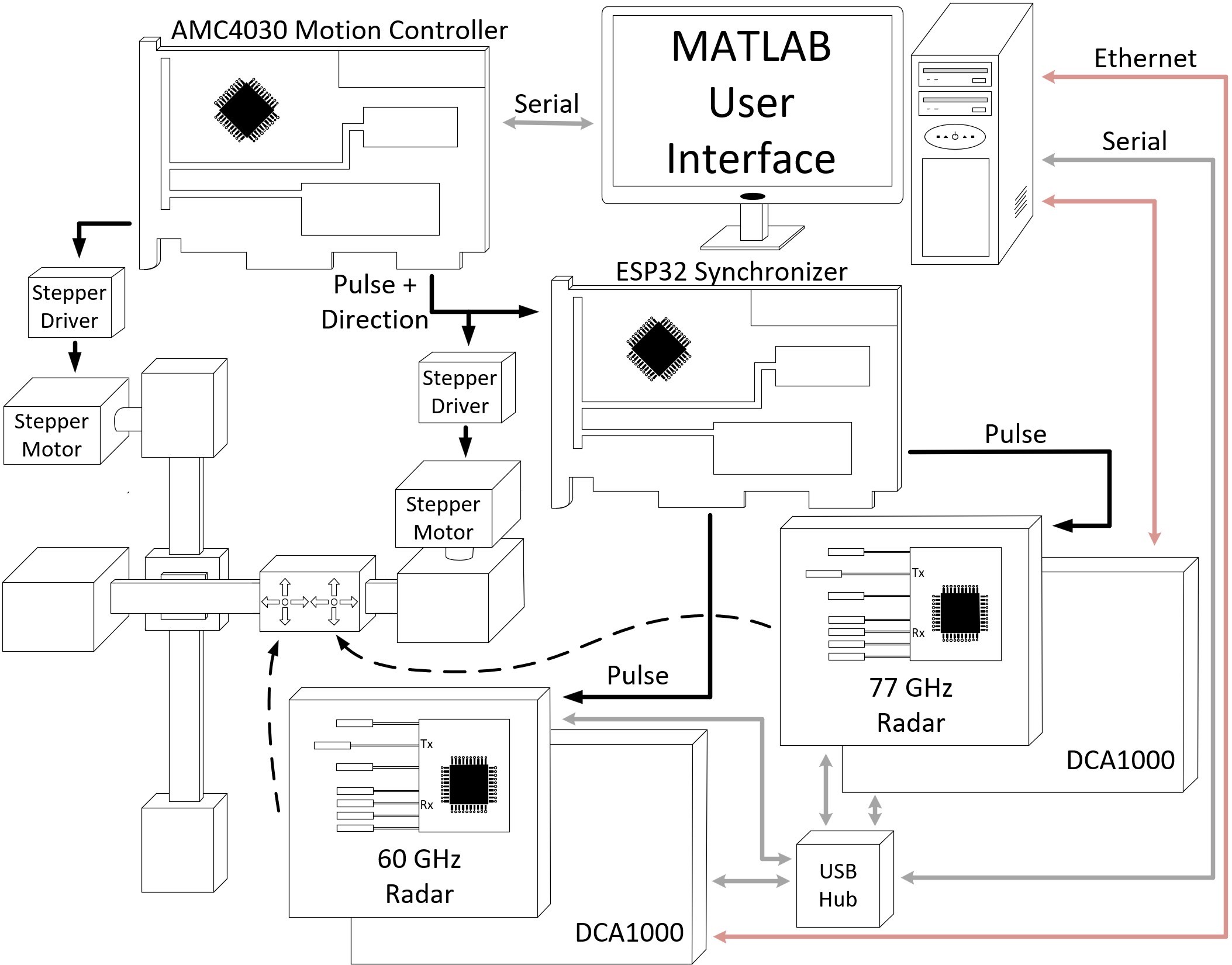}
    \caption{System architecture of the proposed multiband imaging testbed prototype.}
    \label{fig:dri_system} 
\end{figure}

An overview of the system architecture is presented in Fig. \ref{fig:dri_system}. 
The system consists of a 60 GHz radar, 77 GHz radar, two-axis mechanical scanner, motion controller, synchronization module, and a host PC.
The Texas Instruments (TI) IWR6843ISK and IWR1642BOOST are chosen as the single chip 60 GHz and 77 GHz radars, respectively.
Both radars have an operating bandwidth of $B = 4$ GHz; hence, the subbands span 60--64 GHz and 77--81 GHz. 
In addition, the LFM radars are configured using the parameters detailed in Section \ref{subsec:dri_training_details}. 
The data are captured in real-time by the TI DCA1000 evaluation module for each radar and streamed to the host PC over Ethernet. 
Both radars are mounted onto a belt-driven two-axis mechanical scanner, as shown in Fig. \ref{fig:dri_system}, such that the lowest Rx antennas on each radar are aligned, and the radars are separated horizontally by a distance of $\Delta_x$. 

The two-axis mechanical scanner is driven by stepper motors that receive pulses from a motion controller, and the entire system is controlled using a custom MATLAB user interface running on the host PC. 
The radars are scanned in the $x$- and $y$-directions at the spatial sampling Nyquist rate of $\lambda / 4$ in both the horizontal and vertical directions \cite{yanik2019sparse,sheen2001three}. 
Extending the synchronization approach in \cite{yanik2020development}, we design a novel multi-radar synchronizer for the precise positioning of both radars while operating at high scanning speeds. 
The proposed prototype achieves a speed of 500 mm/s, allowing for short scanning times required by applications such as security screening and packaging. 
However, scanning multiple physical radar modules at these speeds while achieving precise synthetic element positioning required to mitigate distortion is challenging and has not been addressed previously in the literature. 
The synchronizer monitors the stepper driver pulses that determine the position of the platform, to which the radars are mounted, as it both accelerates and decelerates. 
Additionally, the synchronizer tracks the positions of each radar and fires them independently to account for the non-uniform timing required to maintain a uniform synthetic aperture and ensure equivalent illumination of the target scene from both radars. 
Additional hardware-specific details of mmWave imaging testbeds can be found in \cite{yanik2020development}.

\section{Experimental Results}
\label{sec:dri_results}
In this section, the superiority of the proposed $kR$-Net is demonstrated using numerical simulations and empirical experiments. 
The matrix Fourier transform (MFT) algorithm \cite{li2008mft} and matrix-pencil algorithm (MPA) \cite{zou2016matrix,wang2018wavenumber} are adopted as comparison baselines for the following experiments. 
After $kR$-Net is trained using the procedure detailed in Section \ref{subsec:dri_training_details}, we conduct experiments on both synthetic and empirical multiband data to validate the performance of $kR$-Net compared with traditional signal processing approaches. 
We consider a dual-band system with the radar signaling parameters discussed previously. 

\subsection{Visual Comparison of Simulation Results}
\label{subsec:dri_qual_sim}
First, we detail various simulation results obtained using the proposed $kR$-Net algorithm for multiband signal fusion. 

\begin{figure}[h]
    \centering
    \includegraphics[width=0.7\textwidth]{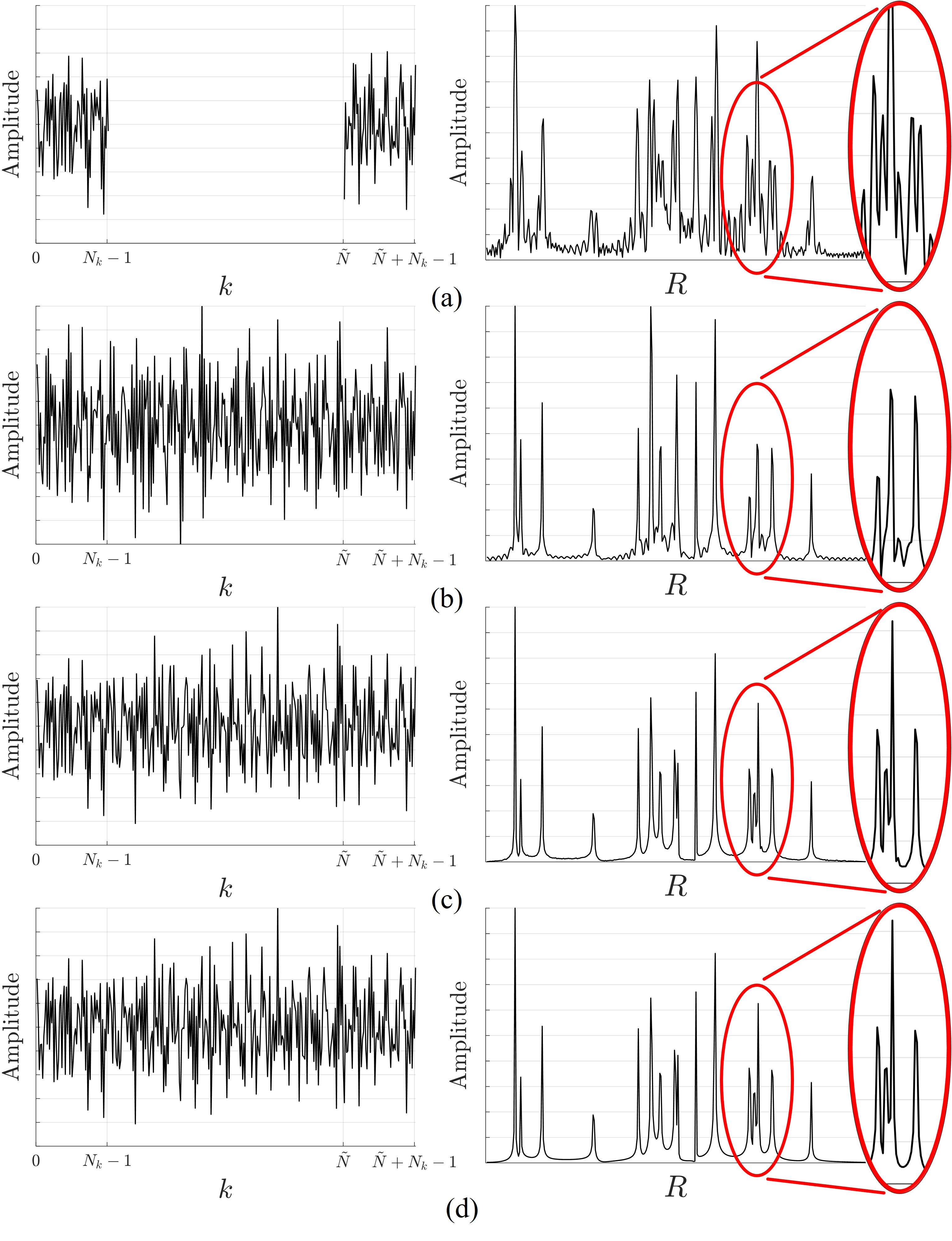}
    \caption{Closely spaced peaks resolved using $kR$-Net. Comparison of multiband signal fusion performance using (a) MFT, (b) MPA, and (c) $kR$-Net compared to (d) the ideal full-band scenario for a single simulated signal consisting of randomly placed point scatters. Left: The real part of the $k$-domain signals. Right: The magnitude of the $R$-domain spectra, demonstrating the super-resolution capability of the proposed $kR$-Net.}
    \label{fig:dri_demo3} 
\end{figure}

\subsubsection{Single Multiband Signal}
\label{subsubsec:demo3}
First, we consider the simple case of a single multiband signal captured at the two subbands used throughout the experiments with a target consisting of randomly placed points. 
The $k$-domain signals and $R$-domain spectra for the MFT, MPA, and $kR$-Net are shown in Fig. \ref{fig:dri_demo3} and compared with the corresponding ideal full-band signal. 
Each subband is corrupted with AWGN at an SNR of 20 dB. 
The MPA imputes the lost signal between the two subbands to recover a signal of length $N$; however, the resulting wideband signal deviates from the ideal signal owing to the assumptions in the MPA. 
Although it outperforms the MFT, the MPA is unable to recover every peak in the $R$-domain, and $kR$-Net yields the most accurate reconstruction of the full-band signal. 

\begin{figure}[th]
    \centering
    \includegraphics[width=0.5\textwidth]{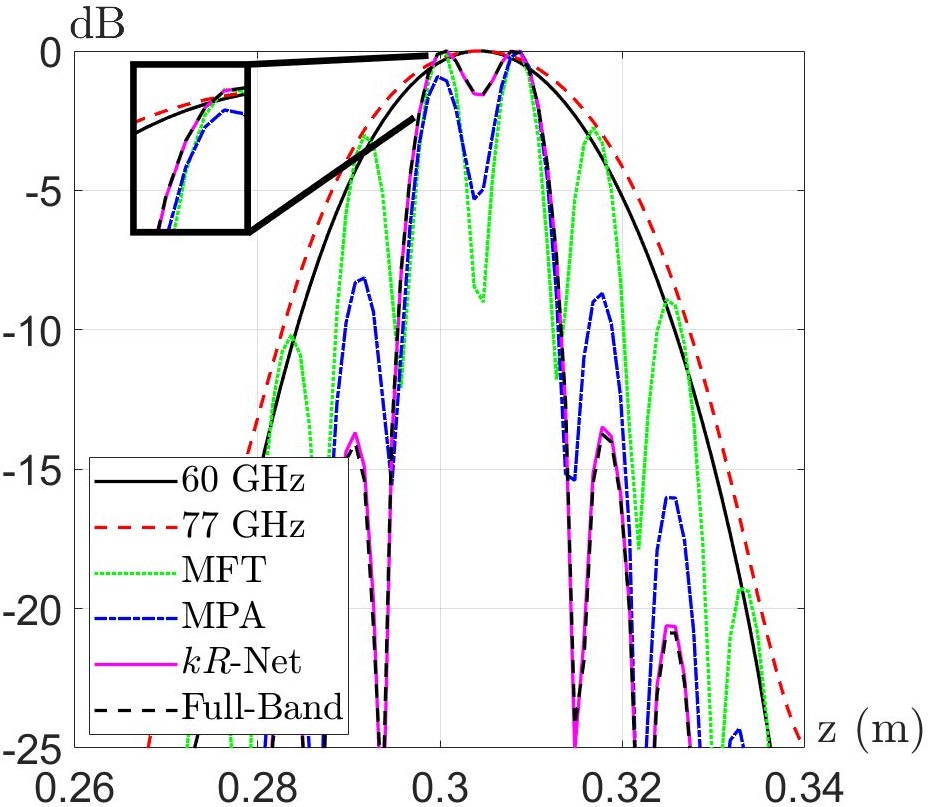}
    \caption{Comparison of imaging results from various scenarios sliced along $y = 0$ m with two simulated point scatterers spaced by $\Delta_z = 7.1$ mm, which corresponds to the minimum resolvable distance for an equivalent bandwidth of 21 GHz.}
    \label{fig:dri_res} 
\end{figure} 

\subsubsection{Effective Bandwidth Study}
\label{subsubsec:dri_res}
Given that the purpose of multiband imaging is to achieve a finer resolution by synthetically increasing the bandwidth, we consider the resolution capability of the proposed algorithm and its corresponding effective bandwidth. 
The resolution of a radar system in the downrange direction is given by $\delta_z = c/2B$, which determines the ability of the system to resolve two closely spaced reflectors. 
Hence, we compare the imaging results for various cases with two closely spaced peaks to evaluate the resolution limit of the algorithms. 
We simulate a scenario with two point scatterers located at $Z_r$ and $Z_r + \Delta_z$ from the radar boresight with AWGN at an SNR of 20 dB. 
Using $Z_r =$ 300 mm, we evaluate the performance for $\Delta_z =$ 7.1 mm, which corresponds to an effective bandwidth of 21 GHz. 
As shown in Fig. \ref{fig:dri_res}, the proposed $kR$-Net achieves a nearly identical response to the ideal full-band signal. 
Since the 60 GHz and 77 GHz radars, subbands 1 and 2, each have a bandwidth of 4 GHz, the two closely spaced reflectors are blurred into a single peak. 
The MFT resolves the two peaks but has severely increased sidelobes compared to the full-band signal.
Because the target consists of two signal components, the MPA resolves both peaks without significant distortion, but demonstrates some minor deviations from the ideal signal.
In contrast, $kR$-Net achieves a more accurate signal with lower sidelobes. 
Thus, the proposed algorithm achieves an effective bandwidth of 21 GHz because the two peaks are clearly resolved. 

\begin{figure*}[th]
    \centering
    \includegraphics[width=\textwidth]{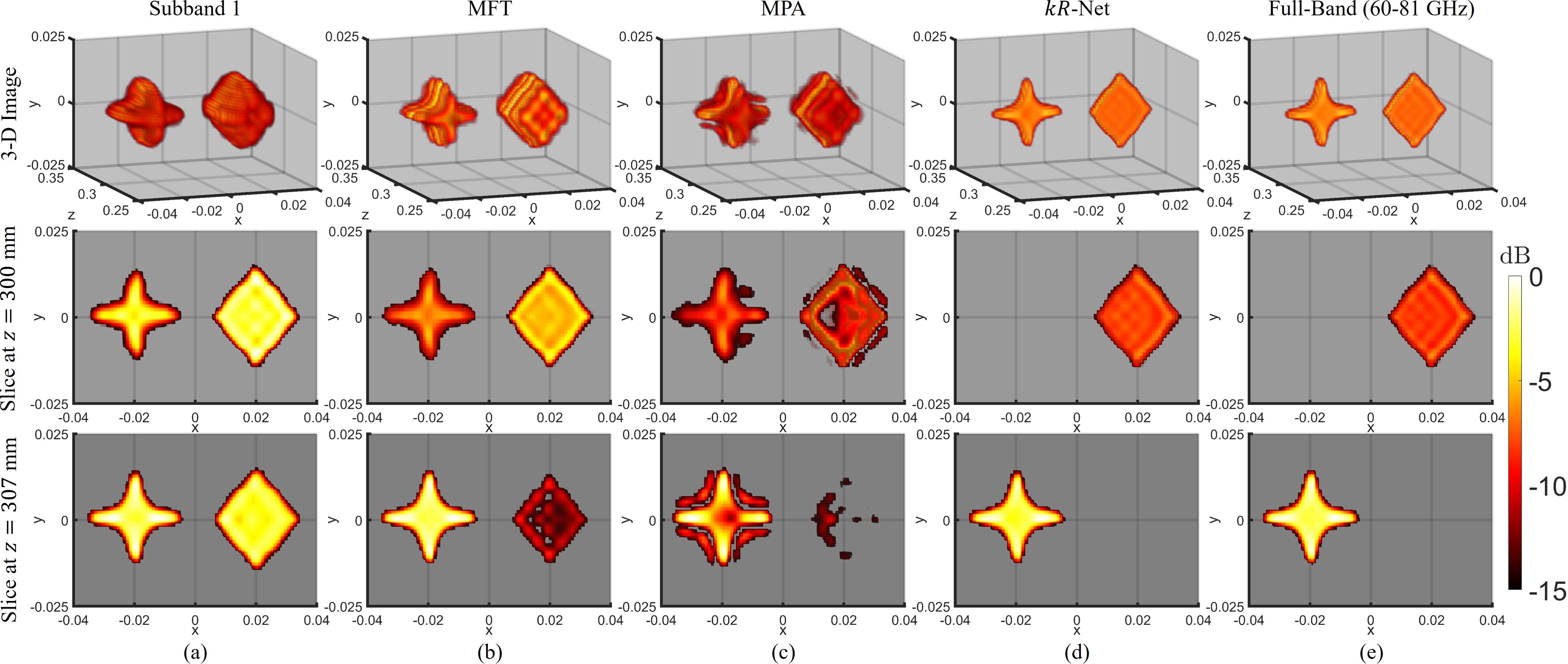}
    \caption{High-bandwidth target consisting of a diamond located at $z = 300$ mm and star located at $z = 307$ mm. The proposed $kR$-Net is able to separate the two shapes to their respective $z$-planes, exhibiting comparable imaging performance to an ideal full-band transceiver without feature loss or visible sidelobes present in the MFT and MPA images. Imaging results for simulated high-bandwidth target using (a) subband 1  (60--64 GHz), (b) MFT, (c) MPA, (d) $kR$-Net, (e) ideal full-band (60--81 GHz). First row: \mbox{3-D} image. Second row: slice at $z = 300$ mm, corresponding to the location of the diamond shape. Third row: slice at $z = 307$ mm, corresponding to the location of the star shape.}
    \label{fig:dri_sim6} 
\end{figure*}

\subsubsection{High-Bandwidth Target}
\label{subsubsec:dri_sim2_cutout1}

To visually evaluate the imaging performance of $kR$-Net and demonstrate the deficiencies of the MFT and MPA on a high-bandwidth target, two shapes are considered under the multiband scenario in the near-field with a planar array with dimensions of 0.125 m $\times$ 0.125 m, satisfying the spatial Nyquist criterion \cite{yanik2019sparse}. 
A star shape is placed on the left at the plane $z = 307$ mm and a diamond to the right at the plane $z = 300$ mm, with the synthetic array at the plane $z = 0$ m. 
For many high-resolution imaging tasks, objects must be localized, classified, or counted to identify concealed weapons, ensure correct packaging, or detect defects. 
To localize closely spaced targets, a transceiver with a bandwidth of 4 GHz may not be adequate as the low range resolution yields images stretched in the $z$-direction beyond the physical dimensions of the objects.
Fig. \ref{fig:dri_sim6} shows the \mbox{3-D} reconstructed images and slices along $z = 300$ mm and $z = 307$ mm to demonstrate the distortion along the $z$-direction that can contaminate the images and degrade system performance.  
As the objects are located in distinct $z$-planes, the second and third rows should show only the diamond and star, respectively, as shown in Fig. \ref{fig:dri_sim6}e, for the ideal full-band case.
The image recovered from subband 1 (60--64 GHz) is shown in Fig. \ref{fig:dri_sim6}a and demonstrates the limitations of low bandwidth, as both shapes are clearly visible in both slices.
Similarly, the MFT results in powerful sidelobes, as shown in Fig. \ref{fig:dri_res}, which correspond to ghost shapes along the $z$-direction for solid targets, as shown in Fig. \ref{fig:dri_sim6}b.
Hence, images recovered using the MFT are not suitable for localization or object counting tasks because they are obscured by spurious sidelobes. 
For this high-bandwidth target scenario, the image recovered using the MPA, shown in Fig. \ref{fig:dri_sim6}c, suffers from a loss of fidelity due to the simplistic multiband fusion model and the assumption of a small number of reflectors. 
As a result, the MPA images are not only contaminated with sidelobes that degrade the slices corresponding to each object but also fail to retain the high-bandwidth features of the objects. 
Comparatively, the images recovered using $kR$-Net, shown in Fig. \ref{fig:dri_sim6}d, closely resemble the ideal full-band images. 
This aligns with the conclusions drawn from the numerical experiments and demonstrates the superior performance of $kR$-Net compared to existing algorithms for realistic high-resolution imaging tasks.

\subsection{Quantitative Investigations}
\label{subsec:dri_quant}
First, we compare the imaging performance of the MFT, MPA, and $kR$-Net numerically, considering a near-field SAR scenario with a planar aperture of $200 \times 200$ synthetic elements. 
It is important to note that the MFT and MPA are classical signal processing approaches, whereas $kR$-Net employs a hybrid approach that combines data-driven techniques with signal processing algorithms. 
Fifty Monte Carlo trials are conducted for each experiment. 
We compare the various algorithms using the structural similarity index measure (SSIM), peak signal-to-noise ratio (PSNR), and normalized root mean square error (NRMSE), where SSIM is defined as
\begin{equation}
    \label{eq:ssim}
    \text{SSIM} = \frac{(2 \mu_\mathbf{x} \mu_\mathbf{y} + C_1)(2 \sigma_{\mathbf{x}\mathbf{y}} + C_2)}{(\mu_\mathbf{x}^2 + \mu_\mathbf{y}^2 + C_1)(\sigma_\mathbf{x}^2 + \sigma_\mathbf{y}^2 + C_2)},
\end{equation}
where $\mathbf{x}$ and $\mathbf{y}$ are the reconstructed image and reference image, respectively; $\mu_\mathbf{x}$, $\sigma_\mathbf{x}$ and $\mu_\mathbf{y}$, $\sigma_\mathbf{y}$ are their corresponding mean values and standard deviations, respectively; and $L$ is the dynamic range of the pixel values.
For stability $C_1 = (k_1 L)^2$ and $C_2 = (k_2 L)^2$ where $k_1 = 0.01$ and $k_2 = 0.03$ by default \cite{wang2021tpssiNet}. 
SSIM quantifies the similarity between $\mathbf{x}$ and $\mathbf{y}$, with a larger value indicating better performance and a maximum value of 1 for a perfect reconstruction. 
Similarly, higher PSNR values (dB) \cite{lim2017enhanced} and lower NRMSE values \cite{wang2021tpssiNet} indicate more accurate image reconstruction.

\begin{table}[h]
\centering
\caption{Comparison of SSIM, PSNR, and NRMSE across different number of targets ($N_t$)  using MFT, MPA, and $kR$-Net}
\label{tab:dri_sim_numerical_Nt}
\resizebox{0.75\textwidth}{!}{%
\begin{tabular}{c||ccc|ccc|ccc}
\multirow{2}{*}{$N_t$} & \multicolumn{3}{c|}{MFT} & \multicolumn{3}{c|}{MPA} & \multicolumn{3}{c}{$kR$-Net} \\ \cline{2-10} 
 
& \multicolumn{1}{c|}{SSIM} & \multicolumn{1}{c|}{PSNR} & NRMSE & \multicolumn{1}{c|}{SSIM} & \multicolumn{1}{c|}{PSNR} & NRMSE & \multicolumn{1}{c|}{SSIM} & \multicolumn{1}{c|}{PSNR} & NRMSE \\ 
\hline \hline

3 & \multicolumn{1}{c|}{0.9970} & \multicolumn{1}{c|}{48.91} & 1.145 & \multicolumn{1}{c|}{0.9997} & \multicolumn{1}{c|}{66.92} & 0.1727 & \multicolumn{1}{c|}{\textbf{0.9999}} & \multicolumn{1}{c|}{\textbf{90.90}} & \textbf{0.01779} \\ 
\hline
10 & \multicolumn{1}{c|}{0.9934} & \multicolumn{1}{c|}{45.81} & 1.142 & \multicolumn{1}{c|}{0.9959} & \multicolumn{1}{c|}{48.03} & 0.7904 & \multicolumn{1}{c|}{\textbf{0.9997}} & \multicolumn{1}{c|}{\textbf{59.36}} & \textbf{0.2439} \\ 
\hline
100 & \multicolumn{1}{c|}{0.9703} & \multicolumn{1}{c|}{39.56} & 1.108 & \multicolumn{1}{c|}{0.9684} & \multicolumn{1}{c|}{39.17} & 1.160 & \multicolumn{1}{c|}{\textbf{0.9809}} & \multicolumn{1}{c|}{\textbf{41.60}} & \textbf{0.8779} \\ 
\hline
400 & \multicolumn{1}{c|}{0.9220} & \multicolumn{1}{c|}{34.07} & 1.097 & \multicolumn{1}{c|}{0.9240} & \multicolumn{1}{c|}{33.56} & 1.164 & \multicolumn{1}{c|}{\textbf{0.9425}} & \multicolumn{1}{c|}{\textbf{35.44}} & \textbf{0.9378} \\ 
\hline
700 & \multicolumn{1}{c|}{0.8989} & \multicolumn{1}{c|}{32.28} & 1.089 & \multicolumn{1}{c|}{0.9061} & \multicolumn{1}{c|}{31.94} & 1.130 & \multicolumn{1}{c|}{\textbf{0.9250}} & \multicolumn{1}{c|}{\textbf{33.72}} & \textbf{0.9208} \\ 
\hline
1000 & \multicolumn{1}{c|}{0.8874} & \multicolumn{1}{c|}{31.39} & 1.068 & \multicolumn{1}{c|}{0.8902} & \multicolumn{1}{c|}{30.82} & 1.143 & \multicolumn{1}{c|}{\textbf{0.9094}} & \multicolumn{1}{c|}{\textbf{32.45}} & \textbf{0.9488} \\ 
\hline
1300 & \multicolumn{1}{c|}{0.8678} & \multicolumn{1}{c|}{30.03} & 1.055 & \multicolumn{1}{c|}{0.8714} & \multicolumn{1}{c|}{29.50} & 1.122 & \multicolumn{1}{c|}{\textbf{0.8922}} & \multicolumn{1}{c|}{\textbf{31.14}} & \textbf{0.9302} \\ 
\hline
Avg. & \multicolumn{1}{c|}{0.9338} & \multicolumn{1}{c|}{37.44} & 1.100 & \multicolumn{1}{c|}{0.9364} & \multicolumn{1}{c|}{40.14} & 0.9644 & \multicolumn{1}{c|}{\textbf{0.9499}} & \multicolumn{1}{c|}{\textbf{46.37}} & \textbf{0.6967} \\ 
\hline
\hline
\end{tabular}%
}
\end{table}

Images are computed using the MFT, MPA, and $kR$-Net for multiband signal fusion of the two subbands, 60--64 GHz and 77-81 GHz, and compared against the image recovered from the ideal full-band scenario spanning the entire bandwidth of 60-81 GHz as in (\ref{eq:fb_sim}). 
First, \mbox{3-D} images are generated with an SNR of 20 dB for the full-band and multiband cases, as discussed in Section \ref{subsec:dri_training_details}. 
The robustness of the various algorithms is compared over a varying number of targets $N_t$, which corresponds to the target bandwidth.
The results are presented in Table \ref{tab:dri_sim_numerical_Nt}. 
The MFT suffers from increased sidelobes because it does not account for the frequency gap. 
As expected, the MPA exhibits tremendous performance for low-bandwidth targets but is plagued by imaging degradation as the target bandwidth and $N_t$ increase. 
In contrast, the proposed $kR$-Net demonstrates robustness compared with classical algorithms, particularly when applied to high-bandwidth targets scenarios. 
This is expected as $kR$-Net is trained on a dataset containing both low- and high-bandwidth targets. 
The best evaluations are shown in boldface in Table \ref{tab:dri_sim_numerical_Nt}, indicating the superiority of $kR$-Net over the MFT and MPA for realistic \mbox{3-D} imaging scenarios. 

\begin{table}[h]
\centering
\caption{Comparison of SSIM, PSNR, and NRMSE across values of SNR using MFT, MPA, and $kR$-Net}
\label{tab:dri_sim_numerical_snr}
\resizebox{0.75\textwidth}{!}{%
\begin{tabular}{c||ccc|ccc|ccc}
\multirow{2}{*}{SNR} & \multicolumn{3}{c|}{MFT} & \multicolumn{3}{c|}{MPA} & \multicolumn{3}{c}{$kR$-Net} \\ \cline{2-10} 
 
& \multicolumn{1}{c|}{SSIM} & \multicolumn{1}{c|}{PSNR} & NRMSE & \multicolumn{1}{c|}{SSIM} & \multicolumn{1}{c|}{PSNR} & NRMSE & \multicolumn{1}{c|}{SSIM} & \multicolumn{1}{c|}{PSNR} & NRMSE \\ 
\hline \hline

20 dB & \multicolumn{1}{c|}{0.95414} & \multicolumn{1}{c|}{33.64} & 1.089 & \multicolumn{1}{c|}{0.9743} & \multicolumn{1}{c|}{39.72} & 0.5510 & \multicolumn{1}{c|}{\textbf{0.9839}} & \multicolumn{1}{c|}{\textbf{42.39}} & \textbf{0.4064} \\ 
\hline
15 dB & \multicolumn{1}{c|}{0.9487} & \multicolumn{1}{c|}{33.08} & 1.153 & \multicolumn{1}{c|}{0.9733} & \multicolumn{1}{c|}{38.72} & 0.6138 & \multicolumn{1}{c|}{\textbf{0.9801}} & \multicolumn{1}{c|}{\textbf{41.06}} & \textbf{0.4680} \\ 
\hline
10 dB & \multicolumn{1}{c|}{0.9465} & \multicolumn{1}{c|}{32.324} & 1.150 & \multicolumn{1}{c|}{0.9560} & \multicolumn{1}{c|}{37.41} & 0.6580 & \multicolumn{1}{c|}{\textbf{0.9816}} & \multicolumn{1}{c|}{\textbf{41.47}} & \textbf{0.4129} \\ 
\hline
5 dB & \multicolumn{1}{c|}{0.9485} & \multicolumn{1}{c|}{34.32} & 1.109 & \multicolumn{1}{c|}{0.9573} & \multicolumn{1}{c|}{37.25} & 0.8203 & \multicolumn{1}{c|}{\textbf{0.9800}} & \multicolumn{1}{c|}{\textbf{41.00}} & \textbf{0.5375} \\ 
\hline
0 dB & \multicolumn{1}{c|}{0.9457} & \multicolumn{1}{c|}{32.48} & 1.131 & \multicolumn{1}{c|}{0.9610} & \multicolumn{1}{c|}{37.61} & 0.6880 & \multicolumn{1}{c|}{\textbf{0.9780}} & \multicolumn{1}{c|}{\textbf{40.76}} & \textbf{0.5009} \\ 
\hline
Avg. & \multicolumn{1}{c|}{0.9487} & \multicolumn{1}{c|}{33.17} & 1.1265 & \multicolumn{1}{c|}{0.9644} & \multicolumn{1}{c|}{38.14} & 0.6662 & \multicolumn{1}{c|}{\textbf{0.9807}} & \multicolumn{1}{c|}{\textbf{41.34}} & \textbf{0.4651} \\ 
\hline
\hline
\end{tabular}%
}
\end{table}

Next, images with 200 randomly distributed point scatterers in addition to one solid object, selected from a set of 10 basic shapes including circle, square, triangle, etc., are generated for performance evaluation.
Table \ref{tab:dri_sim_numerical_snr} presents the results for the different algorithms evaluated with SNR values ranging from 0 dB to 20 dB in increments of 5 dB. 
Because $N_t > 200$ for every image, the targets are considered high-bandwidth, and the MPA is unable to adequately model the target intricacies. 
In this experiment, the MPA outperforms the MFT because of the addition of the solid target, as the MPA slightly reduces the sidelobes from the solid target, which contribute a larger amount of power than the point scatterers. 
However, $kR$-Net demonstrates robustness across low and high SNR, yielding high SSIM and PSNR in conjunction with low NRMSE. 

These analyses validate the superiority of the proposed algorithm for both low- and high-bandwidth target scenarios. 
$kR$-Net overcomes deficiencies in the MPA due to a simplistic model that deems the MPA unsuitable for many near-field imaging applications that require high-resolution imaging of intricate objects without feature loss. 
Furthermore, since $kR$-Net is highly parallelizable, its computation time per image, 12.5 s, is significantly less than that of the iterative MPA \cite{zou2016matrix,wang2018wavenumber}, which requires 1.6 h for \mbox{3-D} imaging. 
As described in Section \ref{subsec:dri_existing_methods}, the same GPU hardware is used to implement all three algorithms. 
An efficient, parallelized implementation of the MPA is employed; however, the computation time remains excessive. 
Alternatively, although the MFT requires only 5.2 s to recover an image \cite{li2008mft}, it does not account for the frequency gap between subbands and consistently demonstrates inferior performance to that of $kR$-Net in terms of image quality.

\subsubsection{Ablation Study}
\label{subsubsec:dri_ablation_study}
To demonstrate the effectiveness of the hybrid, dual-domain approach, we compare two baseline networks with $kR$-Net. 
First, we remove the FFT and IFFT blocks from the network such that it operates exclusively in the $k$-domain. 
In contrast to Fig. \ref{fig:kRNet_overview}a, the signals at the input are in the $k$-domain and remain in the $k$-domain throughout the entire network. 
For imputation problems, such as image completion or multiband signal fusion, leveraging the information positioned throughout the input signal, which in our case is the signal at each subband, to complete missing regions is challenging for a CNN. 
Each pixel of an intermediate representation, that is, the output of a convolution layer, in a CNN depends on a region of the representation in the previous convolution layer. 
Hence, a pixel in a given representation depends on a certain region of the input signal, which is known as the effective receptive field. 
However, the effective receptive field typically grows slowly due to small kernel sizes, implying that the inferred signal in the frequency gap will not be aware of the subband signals until later in the network \cite{iizuka2017globally}. 
This is an issue for multiband fusion because samples between subbands require dependence on subband signals for robust estimation. 
Hence, the performance of wavenumber domain network, called $k$-Net, degrades towards the center of the frequency gap \cite{luo2016understanding} as the information from the subband signals does not adequately impact the prediction of the frequency gap. 

\begin{figure}[h]
    \centering
    \includegraphics[width=0.65\textwidth]{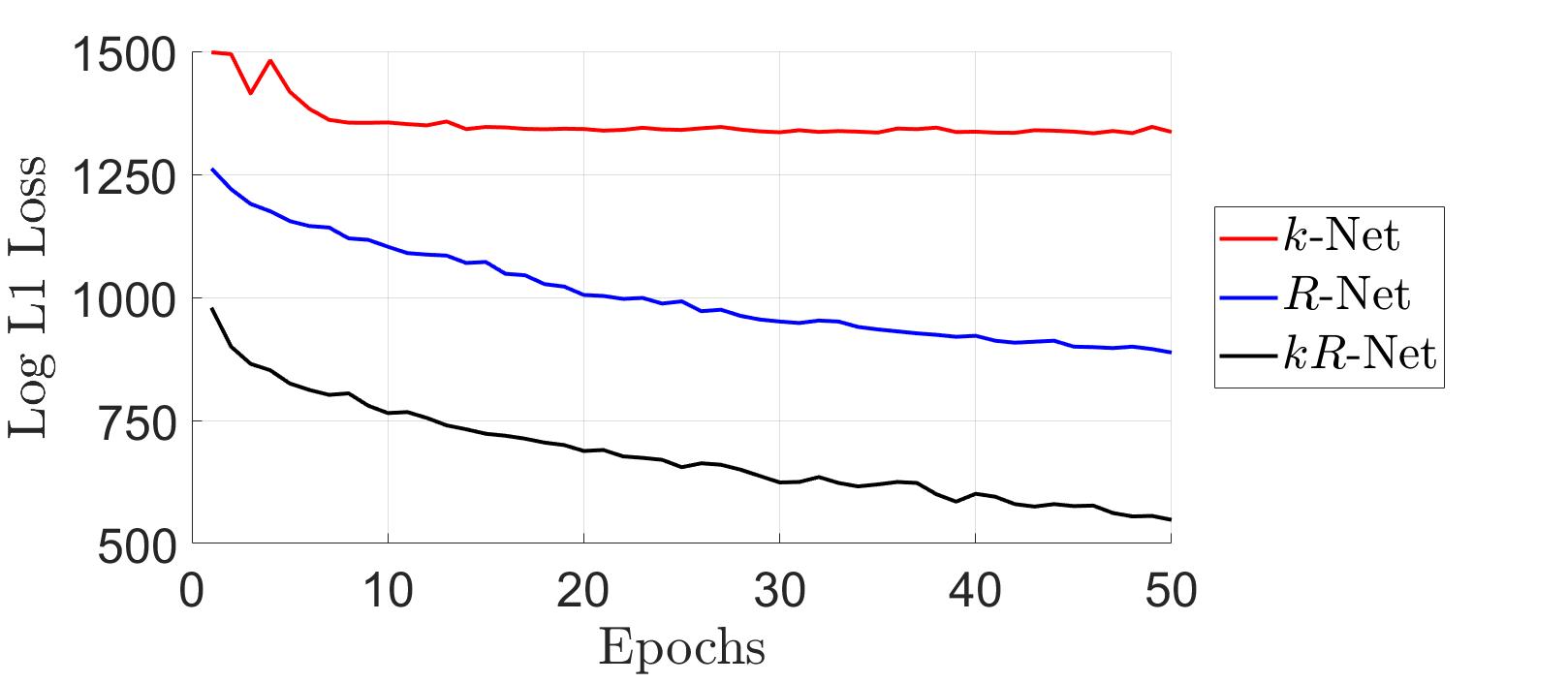}
    \caption{The proposed dual-domain $kR$-Net architecture outperforms conventional CNN models operating on $k$- or $R$-domain signals for training convergence and final performance. Comparison of log-scaled validation loss from 2048 samples during training.}
    \label{fig:dri_convergence} 
\end{figure}

Alternatively, the second baseline employs an architecture identical to that of $k$-Net but performs signal fusion in the $R$-domain. 
This technique, called $R$-Net, represents a spectral super-resolution approach to multiband signal fusion, similar to data-driven line spectra super-resolution algorithms \cite{izacard2021datadriven,pan2021complexFrequencyEstimation}.
However, despite being trained and validated on identical signals, there is a noticeable performance difference as each network operates in a different domain.

First, we compare the convergence of $kR$-Net against the baseline networks $k$-Net and $R$-Net during training, as shown in Fig. \ref{fig:dri_convergence}. 
The three algorithms are trained for 50 epochs on the same dataset, and the same dataset is used for validation.
We observe that the dual-domain architecture of $kR$-Net accelerates network training and improves the final performance. 
Comparatively, $k$-Net converges at a considerably slower rate and the signal fusion performance of $R$-Net is inferior to that of $kR$-Net. 

\begin{table}[h]
\centering
\caption{Comparison of SSIM, PSNR, and NRMSE across different number of targets ($N_t$)  using $k$-Net, $R$-Net, and $kR$-Net}
\label{tab:dri_ablation_Nt}
\resizebox{0.75\textwidth}{!}{%
\begin{tabular}{c||ccc|ccc|ccc}
\multirow{2}{*}{$N_t$} & \multicolumn{3}{c|}{$k$-Net} & \multicolumn{3}{c|}{$R$-Net} & \multicolumn{3}{c}{$kR$-Net} \\ \cline{2-10} 
 
& \multicolumn{1}{c|}{SSIM} & \multicolumn{1}{c|}{PSNR} & NRMSE & \multicolumn{1}{c|}{SSIM} & \multicolumn{1}{c|}{PSNR} & NRMSE & \multicolumn{1}{c|}{SSIM} & \multicolumn{1}{c|}{PSNR} & NRMSE \\ 
\hline \hline

3 & \multicolumn{1}{c|}{0.9972} & \multicolumn{1}{c|}{50.05} & 1.01 & \multicolumn{1}{c|}{0.9995} & \multicolumn{1}{c|}{69.98} & 0.1026 & \multicolumn{1}{c|}{\textbf{0.9999}} & \multicolumn{1}{c|}{\textbf{90.90}} & \textbf{0.01779} \\ 
\hline
10 & \multicolumn{1}{c|}{0.9946} & \multicolumn{1}{c|}{47.15} & 0.9809 & \multicolumn{1}{c|}{0.9994} & \multicolumn{1}{c|}{56.65} & 0.3308 & \multicolumn{1}{c|}{\textbf{0.9997}} & \multicolumn{1}{c|}{\textbf{59.36}} & \textbf{0.2439} \\ 
\hline
100 & \multicolumn{1}{c|}{0.9780} & \multicolumn{1}{c|}{40.92} & 0.9475 & \multicolumn{1}{c|}{0.9780} & \multicolumn{1}{c|}{41.00} & 0.9411 & \multicolumn{1}{c|}{\textbf{0.9809}} & \multicolumn{1}{c|}{\textbf{41.60}} & \textbf{0.8779} \\ 
\hline
400 & \multicolumn{1}{c|}{0.9380} & \multicolumn{1}{c|}{35.13} & 0.9715 & \multicolumn{1}{c|}{0.9351} & \multicolumn{1}{c|}{34.76} & 1.015 & \multicolumn{1}{c|}{\textbf{0.9425}} & \multicolumn{1}{c|}{\textbf{35.44}} & \textbf{0.9378} \\ 
\hline
700 & \multicolumn{1}{c|}{0.9168} & \multicolumn{1}{c|}{33.23} & 0.9746 & \multicolumn{1}{c|}{0.9167} & \multicolumn{1}{c|}{33.05} & 0.9955 & \multicolumn{1}{c|}{\textbf{0.9250}} & \multicolumn{1}{c|}{\textbf{33.72}} & \textbf{0.9208} \\ 
\hline
1000 & \multicolumn{1}{c|}{0.9035} & \multicolumn{1}{c|}{32.17} & 0.9767 & \multicolumn{1}{c|}{0.8999} & \multicolumn{1}{c|}{31.75} & 1.029 & \multicolumn{1}{c|}{\textbf{0.9094}} & \multicolumn{1}{c|}{\textbf{32.45}} & \textbf{0.9488} \\ 
\hline
1300 & \multicolumn{1}{c|}{0.8863} & \multicolumn{1}{c|}{30.96} & 0.9500 & \multicolumn{1}{c|}{0.8816} & \multicolumn{1}{c|}{30.40} & 1.012 & \multicolumn{1}{c|}{\textbf{0.8922}} & \multicolumn{1}{c|}{\textbf{31.14}} & \textbf{0.9302} \\ 
\hline
Avg. & \multicolumn{1}{c|}{0.9449} & \multicolumn{1}{c|}{38.52} & 0.9732 & \multicolumn{1}{c|}{0.9444} & \multicolumn{1}{c|}{42.51} & 0.7751 & \multicolumn{1}{c|}{\textbf{0.9499}} & \multicolumn{1}{c|}{\textbf{46.37}} & \textbf{0.6967} \\ 
\hline
\hline
\end{tabular}%
}
\end{table}

We repeat the experiment comparing performance across different numbers of targets $N_t$, for each of the three networks, and the results are shown in Table \ref{tab:dri_ablation_Nt}, where the best evaluation is marked in boldface.
Although $R$-Net outperforms the MPA in this experiment, it is unable to achieve numerical performance comparable to that of $kR$-Net. 
As $N_t$ increases, the performance of $R$-Net begins to degrade. 
This is likely due to the increased likelihood of closely spaced peaks in the $R$-domain and the resulting spectral blur. 
On the other hand, $kR$-Net demonstrates improved performance compared with both baselines and substantiates the superior ability of a hybrid approach to overcome spectral blur. 
The results demonstrate the superiority of $kR$-Net owing to its hybrid, dual-domain approach and verify that learning in both the $k$- and $R$-domains improves the quantitative performance for multiband signal fusion. 

\subsubsection{Generalizability Study}
\label{subsubsec:dri_generalizability_study}
To investigate the generalizability of the proposed technique, we consider three unique multiband scenarios with different frequency ranges, subband ranges, and number of subbands. 
Since each configuration requires a new network trained on data specific to those subbands, we simulate datasets with the same parameters as detailed in Section \ref{subsubsec:dri_dataset} for each configuration. 
Configuration $\mathcal{A}$ as two subbands with equivalent bandwidths of 2 GHz starting at 30 GHz and 38 GHz, respectively, resulting in a corresponding full-band of 30 GHz to 40 GHz. 
In Configuration $\mathcal{B}$, two subbands are used with different bandwidths, 6 GHz and 10 GHz, operating with starting frequencies of 180 GHz and 210 GHz, respectively, resulting in a full-band equivalent of 180 GHz to 220 GHz. 
Finally, a full-band range of 400 GHz to 460 GHz is achieved in Configuration $\mathcal{C}$ using three subbands with bandwidths of 8 GHz, 10 GHz, and 6 GHz and starting frequencies of 400 GHz, 428 GHz, and 454 GHz, respectively. 
\begin{table}[th]
\centering
\caption{Comparison of average PSNR and imaging time across different multiband configurations results across values of $N_t$ and SNR using MFT, MPA, and the $kR$-Net trained for the respective configuration. }
\label{tab:dri_generalizability_study}
\resizebox{0.75\textwidth}{!}{%
\begin{tabular}{c||ccc|ccc|ccc}
\multirow{2}{*}{ } & \multicolumn{3}{c|}{Configuration $\mathcal{A}$} & \multicolumn{3}{c|}{Configuration $\mathcal{B}$} & \multicolumn{3}{c}{Configuration $\mathcal{C}$} \\ \cline{2-10} 
 
& \multicolumn{1}{c|}{MFT} & \multicolumn{1}{c|}{MPA} & $kR$-Net$_\mathcal{A}$ & \multicolumn{1}{c|}{MFT} & \multicolumn{1}{c|}{MPA} & $kR$-Net$_\mathcal{B}$ & \multicolumn{1}{c|}{MFT} & \multicolumn{1}{c|}{MPA} & $kR$-Net$_\mathcal{C}$ \\ 
\hline \hline

$\bar{N_t}$ & \multicolumn{1}{c|}{31.60} & \multicolumn{1}{c|}{34.12} & \textbf{43.57} & \multicolumn{1}{c|}{31.10} & \multicolumn{1}{c|}{37.41} & \textbf{48.50} & \multicolumn{1}{c|}{32.67} & \multicolumn{1}{c|}{34.07} & \textbf{40.52} \\ 
\hline
$\bar{\text{SNR}}$ & \multicolumn{1}{c|}{33.50} & \multicolumn{1}{c|}{38.82} & \textbf{43.82} & \multicolumn{1}{c|}{33.64} & \multicolumn{1}{c|}{35.18} & \textbf{47.86} & \multicolumn{1}{c|}{33.48} & \multicolumn{1}{c|}{39.19} & \textbf{44.79} \\ 
\hline
\hline
Time (s) & \multicolumn{1}{c|}{\textbf{4.412}} & \multicolumn{1}{c|}{4362} & 9.468 & \multicolumn{1}{c|}{\textbf{4.632}} & \multicolumn{1}{c|}{5276} & 9.673 & \multicolumn{1}{c|}{\textbf{5.086}} & \multicolumn{1}{c|}{8773} & 11.39 \\ 
\hline
\hline
\hline
\end{tabular}%
}
\end{table}

Three unique networks are trained on datasets corresponding to each multiband scenario and named $kR$-Net$_\mathcal{A}$, $kR$-Net$_\mathcal{B}$, $kR$-Net$_\mathcal{C}$. 
To evaluate the numerical performance of these networks, we repeat the experiments by comparing the performance across different numbers of targets $N_t$, and SNR for each configuration. 
The average results across $N_t$ and SNR are shown in Table \ref{tab:dri_generalizability_study} along with the computation time for each algorithm, where the best evaluation is marked in boldface. 
For all three configurations, the proposed approach achieves the best numerical performance, demonstrating the ability of the proposed network to generalize to various realistic conditions. 
In addition, the MFT has the lowest computation time but the least robust image focusing, and the GPU-implemented MPA requires an excess of 1 h to impute the full-band signal with inferior reconstruction to the proposed algorithm. 
Since the MPA is only capable of multiband fusion in the case of two subbands, it must be computed twice for Configuration $\mathcal{C}$ to estimate the signal in the two frequency gaps. 
However, this increases the required computation time for the MPA to 2.4 h for each SAR image for this configuration. 
In contrast, the proposed algorithm only needs to be run once to perform a more robust reconstruction, requiring 11.4 s for Configuration $\mathcal{C}$. 
The proposed $kR$-Net demonstrates generalizability to various multisinusoidal imputation problems across a variety of frequency ranges and subband configurations. 
Provided adequate system design and hardware capable of producing multisinusoidal signals consistent with the pre-designed dataset, our technique can be trained for robust fusion across chip designs, vendors, SAR scanning patterns, etc. 
However, as described in Section \ref{sec:dri_signal_model}, the assumption of frequency-independent scattering properties may not be valid for every application and could limit the robustness of the proposed approach depending on the spectral material characteristics of the expected targets at the operating frequencies. 

\begin{figure}[h]
    \centering
    \includegraphics[width=0.65\textwidth]{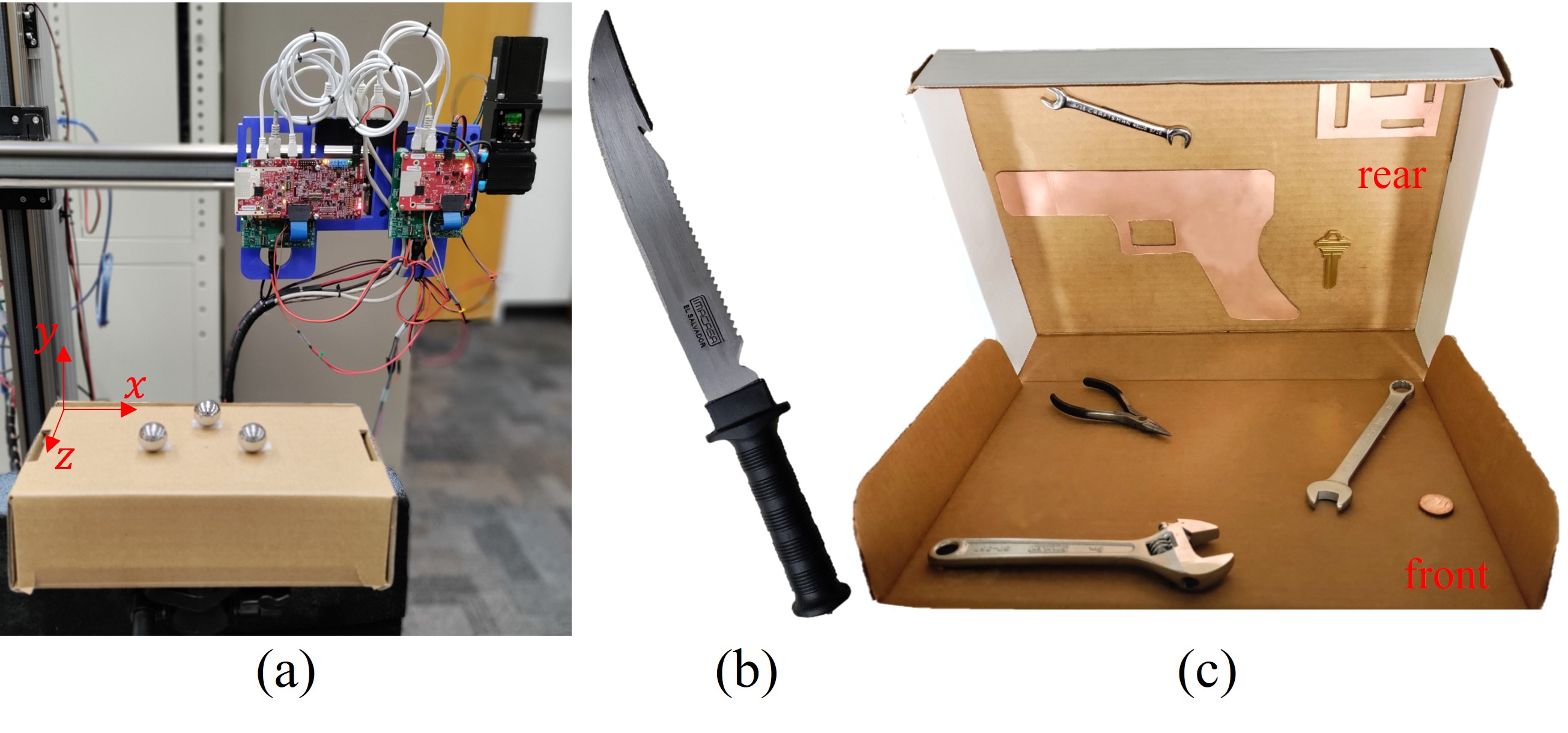}
    \caption{Various experimental targets: (a) metallic sphere targets, (b) large knife with serrated edge and notch near handle, and (c) hidden objects inside a cardboard box.}
    \label{fig:dri_targets} 
\end{figure}

\subsection{Empirical Results}
\label{subsec:dri_qual_exp}
Using the multiband imaging system detailed in Section \ref{sec:dri_system}, we acquire radar data of several objects at the two aforementioned subbands and compare the imaging results of the various multiband fusion algorithms. 

\begin{figure*}[t]
    \centering
    \includegraphics[width=\textwidth]{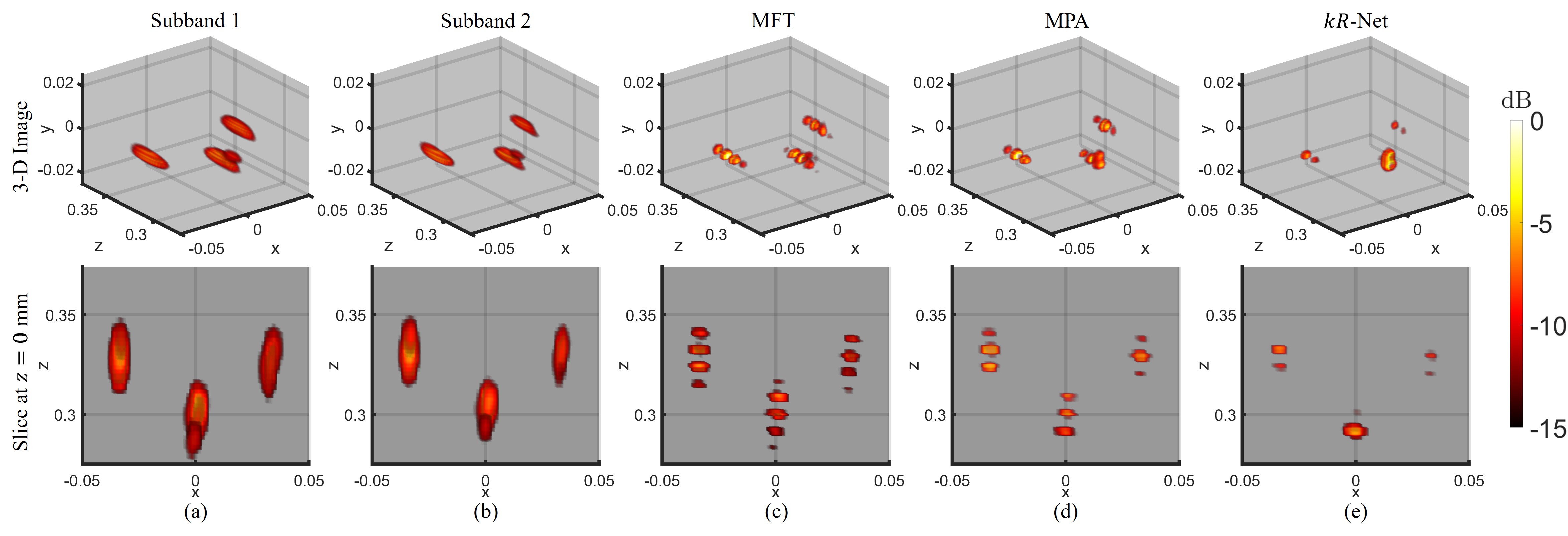}
    \caption{Whereas the MFT and MPA demonstrate ghost artifacts in the range direction, $kR$-Net resolves the three peaks with minimal sidelobes. Imaging results from metallic sphere targets, shown in Fig. \ref{fig:dri_targets}a, using (a) subband 1 (60--64 GHz), (b) subband 2 (77--81 GHz), (c) MFT, (d) MPA, (e) $kR$-Net.}
    \label{fig:dri_exp7} 
\end{figure*}

First, three metallic spheres, each with a diameter of 1.5 cm, are placed in front of the center of the array, as shown in Fig. \ref{fig:dri_targets}a. 
The sphere nearest to the planar scanner is separated by 3 cm from the other two spheres in the $z$-direction. 
The two further spheres are separated by 6 cm such that they are centered around the nearer sphere along the horizontal direction. 
Along the cross-range directions, the images are well resolved and focused; however, the range resolution varies depending on the approach. 
After sampling a planar array with dimensions of 0.125 m $\times$ 0.125 m, the recovered images are computed, as shown in Fig. \ref{fig:dri_exp7}. 
The images recovered from the first and second subbands are shown in Figs. \ref{fig:dri_exp7}a and \ref{fig:dri_exp7}b, respectively, and have low resolution in the $z$-direction because of the bandwidth of 4 GHz. 

Applying the MFT to the collected data yields the image shown in Fig. \ref{fig:dri_exp7}c. 
The MFT image is plagued by sidelobes in the $z$-direction, which obscure the location of each metallic sphere. 
This is due to the fact that the MFT does not account for the missing $k$-domain data in the signal fusion process. 
In contrast, the MPA attempts to fill the frequency gap and is relatively successful because the number of targets is small compared with the number of samples for each subband. 
By applying $kR$-Net, the high-resolution image in Fig. \ref{fig:dri_exp7}e is recovered. 
Compared with the image reconstructed using the MPA, the $kR$-Net image has decreased sidelobes for each of the three spheres. 
Even with a simple target scene, the proposed method demonstrates superior focusing performance compared with the conventional MFT and MPA approaches. 


\begin{figure*}[t]
    \centering
    \includegraphics[width=\textwidth]{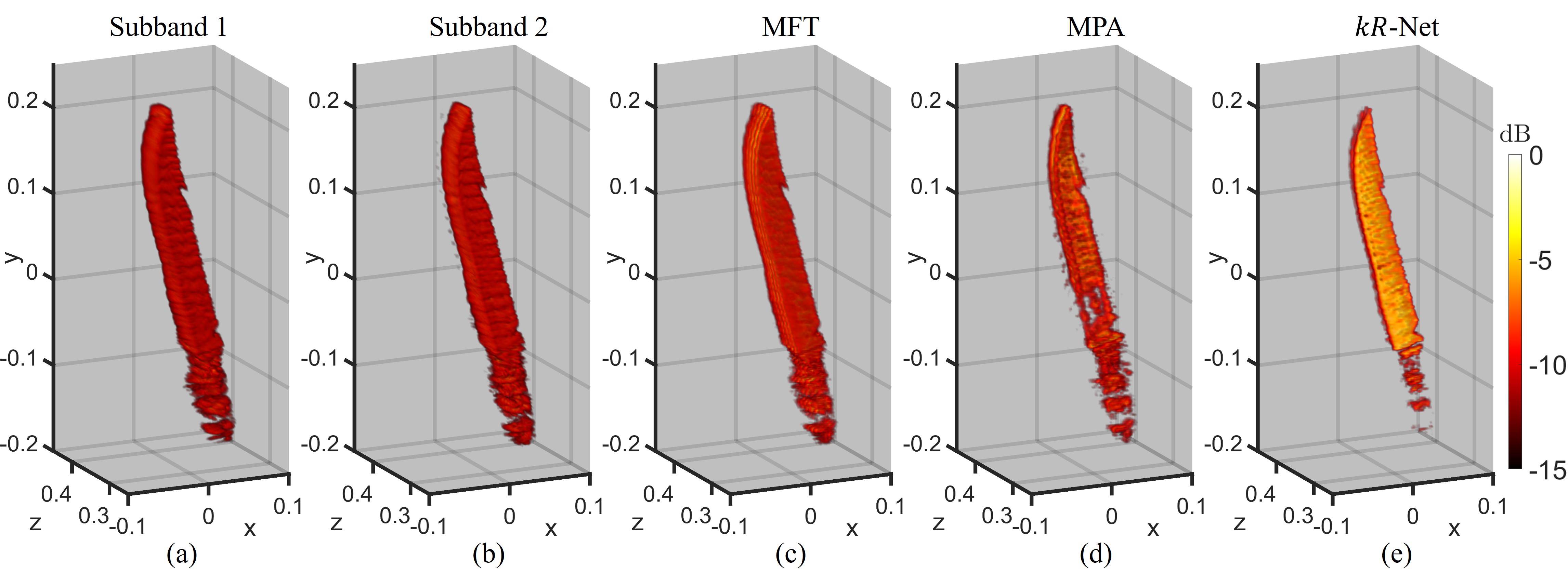}
    \caption{$kR$-Net recovers a high-fidelity image of the knife recovering the thin blade with high depth-resolution, while retaining the shape of the knife, which is lost when applying the MPA. Empirical imaging results from large knife, shown in Fig. \ref{fig:dri_targets}b, using (a) subband 1 (60--64 GHz), (b) subband 2 (77--81 GHz), (c) MFT, (d) MPA, (e) $kR$-Net.}
    \label{fig:dri_exp5} 
\end{figure*} 

Next, we consider a large knife, as shown in Fig. \ref{fig:dri_targets}b. 
A large array is synthesized with dimensions of 0.45 m $\times$ 0.8 m to scan the knife. 
The collected data are processed using the MFT, MPA, and $kR$-Net to improve the resolution, and the results are shown in Fig. \ref{fig:dri_exp5}. 
Again, the images from each subband demonstrate comparable focusing performance but are poorly resolved in the range direction because of the limited bandwidth. 
Without accounting for the missing data in the frequency gap, the MFT image has considerable sidelobes, as shown in Fig. \ref{fig:dri_exp5}c, with a similar appearance to the single-radar images in Figs. \ref{fig:dri_exp5}a and \ref{fig:dri_exp5}b. 
Although the MPA is capable of reducing the sidelobes moderately compared with the MFT, as shown in Fig. \ref{fig:dri_exp5}d, the structure of the knife is not retained because of the simplistic model employed by the MPA. 
As a result, the knife blade is distorted and ghosting is observed along the range direction. 
For concealed weapon detection or occluded item recognition, the poor reconstruction quality of the MPA for high-bandwidth targets, such as this knife, may prohibitively degrade the system performance. 
However, $kR$-Net demonstrates the best focusing performance by achieving a fine resolution in the $z$-direction while retaining the intricate features of the target. 
The serrated edge and notch on the knife are clearly visible in the recovered image shown in Fig. \ref{fig:dri_exp5}e, and the handle closely resembles the physical dimensions shown in Fig. \ref{fig:dri_targets}b. 

\begin{figure*}[t]
    \centering
    \includegraphics[width=\textwidth]{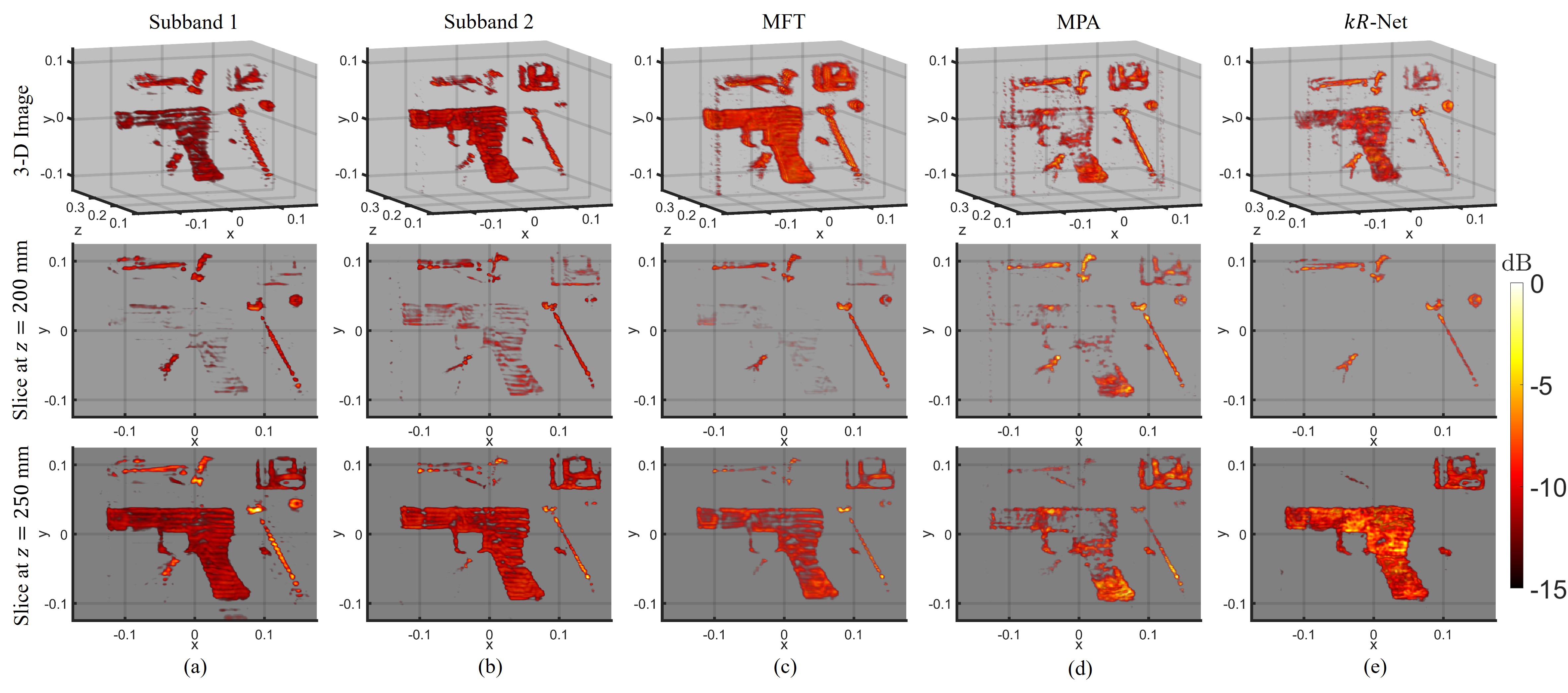}
    \caption{The proposed $kR$-Net separates the items at the front and rear of the box demonstrating high-fidelity super-resolution for a realistic hidden item scenario and superior performance compared with the MFT and MPA. 
    Empirical imaging results from hidden objects target, shown in Fig. \ref{fig:dri_targets}c, using (a) subband 1 (60--64 GHz), (b) subband 2 (77--81 GHz), (c) MFT, (d) MPA, (e) $kR$-Net. First row: \mbox{3-D} image. Second row: slice at $z = 200$ mm, corresponding to the location of the front of the box. Third row: slice at $z = 250$ mm, corresponding to the location of the rear of the box.}
    \label{fig:dri_exp11} 
\end{figure*}

Finally, we consider a hidden object scenario with several items inside a cardboard box, as shown in Fig. \ref{fig:dri_targets}c. 
The box is positioned such that the front of the box is located 200 mm from the radar boresight and parallel to the synthetic array. 
The items attached to the front of the box are separated from those in the rear by 5 cm and the box is illuminated by an array with dimensions of 0.25 m $\times$ 0.125 m. 
For more a closer spacing along the $z$-direction, the imaging results are expected to be further degraded. 
Fig. \ref{fig:dri_exp11} shows the reconstructed \mbox{3-D} images and slices at $z = 200$ mm and $z = 250$ mm, corresponding to the front and rear of the box, respectively. 
For accurate high-resolution imaging, the slice at $z = 200$ mm should contain only the objects at the front of the box and the slice at $z = 250$ mm should contain only the objects at the rear of the box. 

For the two subbands with bandwidths of 4 GHz, the recovered images are spread across the $z$-direction, and both slices shown are contaminated by objects from the front and rear, as shown in Figs. \ref{fig:dri_exp11}a and \ref{fig:dri_exp11}b. 
The small wrench and key, both located at the rear of the box, suffer from weaker reflections and occlusion and are not well resolved by the algorithms tested. 
A cylindrical SAR approach \cite{smith2020nearfieldisar,gao2016efficient} or image enhancement algorithm  \cite{dai2021imaging,jing2022enhanced,smith2021An,vasileiou2022efficient,zhang2019target,smith2022ffh_vit,yin2021study} may improve image quality in the case of occlusion. 
The image recovered using the MFT, shown in Fig. \ref{fig:dri_exp11}c, exhibits the expected behavior, as sidelobes along the range direction cause ghosting, which results in the objects being visible in both range slices. 
Similarly, the MPA reduces the sidelobes moderately compared to the MFT, as shown in Fig. \ref{fig:dri_exp11}d. 
However, as expected from prior experiments and the inherent limitations of the MPA, the sidelobes are not mitigated, and some features of the objects are lost. 
The image recovered using $kR$-Net is shown in Fig. \ref{fig:dri_exp11}e and demonstrates improved performance in two key respects. 
First, the $kR$-Net image retains the high-fidelity features of the target, which are necessary for a host of applications, including image segmentation and object classification. 
The image quality of the wrenches is particularly notable, as the jaw of each wrench is more clearly visible compared with the images recovered from the existing approaches. 
Secondly, the ghosting along the $z$-direction is significantly reduced, and the objects at the front side of the box are visible only in the $z = 200$ mm slice. 
Likewise, the objects at the rear side of the box are only visible in the $z = 250$ mm slice. 
Without the contamination observed using the MFT or MPA, the objects can be more easily localized and classified, enabling super-resolution for a host of imaging applications.  
The proposed hybrid, dual-domain algorithm yields high-resolution, high-fidelity images without feature loss and demonstrates improved performance over existing techniques in realistic scenarios. 

Through numerical simulation and empirical analysis, the proposed algorithm demonstrates superior performance to the MFT and MPA in terms of efficiency and image quality.
The proposed $kR$-Net offers improved robustness for low- and high-bandwidth target scenarios in addition to low SNR conditions. 
For practical imaging of complex, sophisticated targets, $kR$-Net achieves spatial super-resolution by improved multiband signal fusion without compromising the intricate features of the target.
Hence, $kR$-Net is better suited for high-resolution multiband imaging applications. 

\section{Conclusion}
\label{sec:dri_conclusion}
In this article, we introduce a novel deep learning-based algorithm for multiband signal fusion for \mbox{3-D} SAR super-resolution. 
By approaching the signal fusion problem from the wavenumber domain, we observe that imputation in the $k$-domain signal is equivalent to super-resolution in the wavenumber spectral domain. 
Hence, the proposed network employs a hybrid, dual-domain residual architecture that leverages the relationships in the $k$-domain and $R$-domain for improved performance. 
We develop a novel residual CV-CNN framework with domain transformation blocks interspersed throughout the network, resulting in superior performance compared with conventional CNN models. 
Compared with the MPA, which assumes a small number of reflectors in the scene or a low-bandwidth target, the proposed $kR$-Net is robust for imaging scenarios containing intricate targets consisting of many reflectors. 
Through simulation and empirical validation, $kR$-Net demonstrates superior imaging performance for multiband signal fusion for both low- and high-bandwidth targets. 
The hybrid architecture outperforms the equivalent single-domain networks operating in either the $k$-domain or $R$-domain. 
Extensive numerical investigations validate the superiority of $kR$-Net compared with the conventional MFT and MPA for signal fusion, in addition to single-domain CNN models. 
Using a custom multi-radar mechanical prototype built from commercially available mmWave radars, we conduct imaging experiments on various targets and observe significantly improved performance of $kR$-Net in terms of image focusing and efficiency. 

\bibliography{mega_bib}
\bibliographystyle{IEEEtran}

\end{document}